\documentclass[11pt]{article}

\usepackage[preprint]{acl}
\usepackage{times}
\usepackage{latexsym}
\usepackage{booktabs}

\usepackage[T1]{fontenc}
\usepackage[utf8]{inputenc}
\usepackage{microtype}

\usepackage{inconsolata}

\usepackage{graphicx}

\usepackage{amsmath}
\usepackage{amssymb} 
\usepackage{tabularx,makecell,booktabs,array,multirow}
\newcolumntype{Y}{>{\raggedright\arraybackslash}X}

\usepackage[most]{tcolorbox}
\usepackage{multirow}
\usepackage{colortbl}
\usepackage[table]{xcolor}
\usepackage{adjustbox}
\definecolor{mutedblue}{HTML}{4F81BD}

\newtcolorbox{promptbox}[2][]{%
  colback=white,
  colframe=black,
  colbacktitle=gray!15,
  coltitle=black,
  fonttitle=\bfseries\centering,
  title={#2},
  sharp corners,
  boxrule=0.8pt,
  left=3pt, right=3pt, top=3pt, bottom=3pt, 
  #1
}
\usepackage{color}

\definecolor{wrongred}{RGB}{200, 0, 0}
\definecolor{correctblue}{RGB}{0, 100, 180}

\usepackage{amsthm}
\usepackage{thmtools}
\theoremstyle{plain}
\newtheorem{definition}{Definition}[section]

\usepackage{enumitem}
\usepackage{subcaption}

\usepackage{xcolor}
\usepackage{tabularx}
\usepackage{array}

\newcolumntype{Y}{>{\hsize=2\hsize}X}

\usepackage{lipsum} % Load the lipsum package

\usepackage{hyperref}

\title{Concretized Proposition Prompting Resolves Composition-Knowledge Dichotomy in Large Language Models}

\author{
  Changhun Lee$^{1,2}$\thanks{\ \ Equal contribution}\thanks{\ \ Corresponding author},
  Minguk Jeon$^{2}$\footnotemark[1],
  Jongkyung Shin$^{2}$ \and
  Chiehyeon Lim$^{2, 3}$
  \\
  \texttt{cl4670@cumc.columbia.edu}$^{1}$, 
  \texttt{chlim@posco-inc.com}$^{3}$\\
  \texttt{\{changhun, rzbsys, shinjk1156, chlim\}}@unist.ac.kr$^{2}$
  \\
  $^{1}$Columbia University \quad
  $^{2}$UNIST \quad
  $^{3}$POSCO Holdings
}

\begin{document}
\maketitle
\begin{abstract}
% \lipsum[1] 
LLMs often struggle to balance compositionality with knowledgeability, a challenge we define as Composition-Knowledge Dichotomy. To address this, we propose Concretized Proposition Prompting (CPP), a framework that explicitly concretizes propositions relevant to questions. The results demonstrate that CPP significantly enhances reasoning performance, particularly in medical benchmarks where precise knowledge is paramount, while being competitive on math benchmarks where deductive reasoning is prioritized. Additional experiments reveal that CPP is scalable to various foundation models and parameter sizes, being a fundamental paradigm that bridges the gap between composition- and knowledge-based approaches. Consequently, CPP resolves the composition-knowledge dichotomy by providing a solid foundation for logically organized and factually grounded reasoning.
\end{abstract}

% and provides a solid foundation for logically organized and factually grounded reasoning.

\section{Introduction} \label{sec: introduction}

Despite the great success of large language models (LLMs) \citep{achiam2023gpt, touvron2023llama, comanici2025gemini}, there still remains a gap between human intelligence and LLMs in terms of their reasoning capabilities. The recently proposed \textit{chain-of-thought} (CoT) prompting methods \citep{wei2022, chowdhery2022, kojima2022} have significantly narrowed this gap. The core idea of CoT prompting is to combine rationale generation \citep{ling-etal-2017-program} with few-shot prompting \citep{brown2020} in a way that so-called `thoughts' are provided to LLMs in the format of \texttt{<input, thoughts, output>}. 

Though CoT prompting can improve the reasoning capabilities of LLMs \citep{zhou2022, xia-etal-2025-beyond}, it still has critical limitations. For example, it cannot match human performance at commonsense and multi-step reasoning \citep{sprague2024}, is not as useful as in areas beyond math \citep{kambhampati2024}, can even hurt performance \citep{NEURIPS2024_ad236edc, Nakkiran2025TrainedOT}, and incur computational costs more than the performance gains it delivers \citep{sprague2025to}. Furthermore, since the CoT is generated by LLMs that are vulnerable to hallucination, it has a risk of inducing erroneous post-hoc rationalization (i.e., exquisite hallucination) for the LLMs taking those hallucinated CoT to generate incorrect answers \citep{huang2025survey, cheng-etal-2025-chain, lewis-lim-etal-2025-analysing, arcuschin2025chain}.

In response to these limitations, advanced CoT techniques have been suggested. One main approach focuses on organizing the structure of LLM reasoning. For example, \textit{Least-to-Most} prompting breaks down a complex problem into simpler subproblems \cite{zhou2022}, \textit{Plan-and-Solve} devises a plan first and solves a problem next \cite{wang-etal-2023-plan}, \textit{Thread of Thought} segments chaotic contexts into manageable parts \cite{zhou2023threadthoughtunravelingchaotic}, \textit{Meta-Prompting} prioritizes the structured template for how to think over specific examples of what to think \cite{zhang2023meta}. Another approach focuses on retrieving evidence useful for LLM reasoning. \textit{Analogical} prompting generates problem-specific exemplars or knowledge before solving problems \cite{yasunaga2023}, \textit{Self-Knowledge Explicitation (SKE)} explicitly generates verifiable knowledge \cite{huang2025}, and \textit{System-2-Attention} regenerates the problem-relevant contexts \cite{weston20232attentionisneed}.

Both approaches have successfully enhanced the reasoning capabilities of LLMs. However, they are largely polarized into two divergent directions---\textit{compositionality} and \textit{knowledgeability}---which prioritize reasoning structures and grounded evidence, respectively. The former is vulnerable to exquisite hallucinations due to the lack of truth-value discernment. Therefore, it struggles with clinical problems that necessitate precise knowledge for accurate reasoning. Conversely, the latter is vulnerable to erratic deductions due to an absence of logical modes. Hence, it suffers from mathematical problems that require predicate logic to build a chain of reasoning. To bridge this gap, we propose Concretized Proposition Prompting (CPP), a framework designed to integrate these dual axes, where compositionality represents the logical mode of statement, while knowledgeability refers to the truth value of the statement.

% while the latter lacks logical modes, leading to difficulties in mathematical problems that require predicate logic for reasoning. To overcome these limitations by integrating both perspectives, we propose a novel prompting method called \textbf{Concretized Proposition Prompting (CPP)} designed to concretize relevant propositions along with the dual axes of compositionality and knowledgeability. Here, the compositionality represents the logical mode of affirmation or negation, while the knowledgeability refers to the truth value of information.

In this study, we demonstrate that CPP achieves superior performance across question-answer (QA) benchmarks compared to other prompting methods solely focusing on either compositionality or knowledgeability. The experiment includes a total of eight datasets spanning three QA domains: commonsense, math, and medicine. The contributions of our work are as follows:
\begin{itemize}
    \item We propose a simple yet powerful method called Concretized Proposition Prompting (CPP) that is designed to surface relevant propositions by integrating the dual axes of \textit{compositionality} and \textit{knowledgeability}.

    \item We demonstrate that CPP outperforms or competes with other prompting methods across multiple QA datasets, including commonsense, math, and medical benchmarks.

    \item We show that CPP performs consistently across various foundation models, including \texttt{Llama}, \texttt{Qwen}, \texttt{Phi}, \texttt{Gemma}, and \texttt{Mistral}, and scales to model sizes ranging from 7B to 72B.
\end{itemize}

\section{Preliminaries}
In this work, we consider QA tasks where LLMs generate a response $\hat{y}$ to answer a question $q$. The dataset consists of ground-truth question-answer pairs, $\mathcal{D}=\{(q_{i}, a_{i}) \}_{i=1}^{N}$, where $a_{i}$ denotes an answer to the $i$-th question $q_{i}$. 

\subsection{Prompting-based QA task}
Large language models build the probability distribution of next tokens, $\mathcal{M}(y) = \prod_{t=1}^{T} P(y_{t}|y_{t-1})$, and generate a full sentence by sampling tokens autoregressively, $\hat{y}_{1:T} \sim \mathcal{M}(y)$. In the context of QA tasks, the probability distribution is extended to the conditional distribution, $\mathcal{M}_{\phi}(y|q) = \prod_{t=1}^{T}P(y_t|y_{t-1}, q, \phi)$, where $q$ is a question and $\phi$ is a prompt. The prompt contains an instruction, $\mathcal{I}_{\text{task}}$, that describes the target task. QA task is then framed as a prompt-based generative process:
\begin{align*}
    \hat{a} =  \operatorname{extract}(\hat{y}_{1:T}) \quad \text{where} \quad \hat{y}_{1:T} \sim \mathcal{M}(\cdot|\mathcal{I}_{\text{task}}, q) 
\end{align*}
Here, $\operatorname{extract}(\cdot)$ is a parsing function that extracts answer-relevant content from $\hat{y}_{1:T}$.
% \footnote{Note that subjective and objective questions have different answer formats. Hence, the extraction method must be adjusted accordingly.}

\subsection{Chain-of-Thought Prompting}
Chain-of-Thought (CoT) prompting is an emerging technique to enhance the reasoning capabilities of LLMs \citep{wei2022, chowdhery2022, kojima2022} by leveraging a formatted prompt of \texttt{<input, thoughts, output>}. Here, `thoughts' refers to the intermediate reasoning steps self-generated by LLMs \citep{xia-etal-2025-beyond}. Specifically, LLMs equipped with CoT prompting generate a sequence of tokens, with $K$ reasoning steps, $\hat{y}_{1:t-1}$, followed by a response $\hat{y}_{t:T}$:
\begin{align*}
    \hat{y}_{1:T} = \big[ \underbrace{\hat{r}^{(1)}, \cdots, \hat{r}^{(K)}}_{\ = \ \hat{y}_{1:t-1}}, \hat{y}_{t:T} \big] \sim \mathcal{M}(\cdot|\mathcal{I}_{\text{task}}, \mathcal{I}_{\text{cot}}, q)
\end{align*}
where $\mathcal{I}_{\text{cot}}$ is the CoT instruction generally given by \textit{``Let's think step by step''} \citep{kojima2022}. This instruction elicits the model to generate the chain of thoughts (or rationales) and final response. CoT prompting not only enhances the model's capability to solve complex tasks \citep{wang2022, lyu-etal-2023-faithful} but also functions as an interpretable window providing transparency for the model's reasoning process \citep{wei2022, yu2023towards}.

\subsection{DSPy-based Prompt Optimization}
\label{sec:dspy_optimization}
Unlike standard prompting approaches that rely on manual prompt engineering, DSPy automatically optimizes prompts \citep{khattab2023, opsahl-ong-etal-2024-optimizing}. Central to the DSPy framework is the parameterized prompt $\phi$, a series of instructions (e.g., $\phi=[\mathcal{I}_{\text{task}}; \mathcal{I}_{\text{cot}}]$), and optimizes it with respect to a validation metric $\mathcal{R}$. The optimization process is executed by \textit{``Optimizer,''} whose objective is to find the optimal prompt $\phi^*$ that maximizes the expected value of the validation metric over the data distribution $(q, a) \sim \mathcal{D}$,
\begin{align*}
    &\phi^{*} = \operatorname*{argmax}_{\phi} \mathbb{E}_{\substack{(q,a)\sim\mathcal{D} \\ \hat{y}\sim\mathcal{M}_{\phi}(\cdot\mid q)}}
    \left[ \mathcal{R}(\operatorname{extract}(\hat{y}), a) \right] \ ,
\end{align*}
where $\hat{y} \sim \mathcal{M}_{\phi}(\cdot| q)$ is the model output,
$\operatorname{extract}(\hat{y})$ indicates the parsed answer $\hat{a}$, and
$\mathcal{R}(\hat{a}, a) \triangleq \mathbb{I}[\hat{a}=a]$ is an indicator function that returns binary values: one if the answer is correct, zero otherwise. Consequently, we obtain the optimized model $\mathcal{M}_{\phi^*}$ that generates correct answers.

% execute the intended operation. 

% the \textit{Signature} (e.g., \texttt{"question $\to$ answer"}), a declaration that abstractly defines the input and output fields. Based on this signature, a parameterized prompt $\phi$ is optimized to execute the intended operation. Here, 

% The optimization process is governed by a \textit{Teleprompter}.\footnote{According to a \href{https://dspy.ai/learn/optimization/optimizers/?h=optimizer}{DSPy document}, the term \textit{``Teleprompter''} has recently been updated to \textit{``Optimizer.''}
% } Specifically, given an LLM equipped with the parameterized prompt, $\mathcal{M}_{\phi}$, the teleprompter aims to 

\begin{table*}[t]
\small
\centering
\resizebox{\textwidth}{!}{% <-- 페이지 폭에 맞춤
\begin{tabular}{@{}cccll@{}}
\toprule
\begin{tabular}[c]{@{}c@{}}Category\end{tabular}                     & \begin{tabular}[c]{@{}c@{}}Symbolic \\ Representation\end{tabular} & \begin{tabular}[c]{@{}c@{}}Operational\\ Definition\end{tabular} & \multicolumn{1}{c}{Examples of Concretized Propositions} & \multicolumn{1}{c}{QA Domain} \\ \midrule
\multirow{3}{*}{\begin{tabular}[c]{@{}c@{}}True-Positive\\ (TP)\end{tabular}}  & \multirow{3}{*}{($+, 1$)}                                    & \multirow{3}{*}{Affirming a Fact}                                & \textit{New York is a city in the United States}              & Commonsense                   \\ \cmidrule(l){4-5} 
                                                                               &                                                              &                                                                  & \textit{The sum of interior angles of a triangle is 180$^{\circ}$}     & Math                          \\ \cmidrule(l){4-5} 
                                                                               &                                                              &                                                                  & \textit{Antibiotics treat strep throat}                       & Medicine                   \\ \midrule
\multirow{3}{*}{\begin{tabular}[c]{@{}c@{}}True-Negative\\ (TN)\end{tabular}}  & \multirow{3}{*}{($-, 0$)}                                    & \multirow{3}{*}{Negating a Fallacy}                              & \textit{Paris is NOT a city in the United States}             & Commonsense                   \\ \cmidrule(l){4-5} 
                                                                               &                                                              &                                                                  & \textit{The sum of interior angles of a triangle is NOT 200$^{\circ}$} & Math                          \\ \cmidrule(l){4-5} 
                                                                               &                                                              &                                                                  & \textit{Antibiotics do NOT treat the common cold}             & Medicine                   \\ \midrule
\multirow{3}{*}{\begin{tabular}[c]{@{}c@{}}False-Positive\\ (FP)\end{tabular}} & \multirow{3}{*}{($+, 0$)}                                    & \multirow{3}{*}{Affirming a Fallacy}                             & \textit{Paris is a city in the United States}                 & Commonsense                   \\ \cmidrule(l){4-5} 
                                                                               &                                                              &                                                                  & \textit{The sum of interior angles of a triangle is 200$^{\circ}$}     & Math                          \\ \cmidrule(l){4-5} 
                                                                               &                                                              &                                                                  & \textit{Antibiotics treat the common cold}                    & Medicine                   \\ \midrule
\multirow{3}{*}{\begin{tabular}[c]{@{}c@{}}False-Negative\\ (FN)\end{tabular}} & \multirow{3}{*}{($-, 1$)}                                    & \multirow{3}{*}{Negating a Fact}                                 & \textit{New York is NOT a city in the United States}          & Commonsense                   \\ \cmidrule(l){4-5} 
                                                                               &                                                              &                                                                  & \textit{The sum of interior angles of a triangle is NOT 180$^{\circ}$} & Math                          \\ \cmidrule(l){4-5} 
                                                                               &                                                              &                                                                  & \textit{Antibiotics do NOT treat strep throat}                & Medicine                   \\ \bottomrule
\end{tabular}
}
\caption{\textbf{Proposition Categories and Corresponding Examples.} Based on proposition definitions (e.g., \textit{``Affirming a fact.''}), the question-relevant propositions are concretized (e.g., \textit{``New York is a city in the United States.''}) by the proposition model and delivered to the answer model, which makes an answer grounded in concrete propositions.}
\label{tab1: cpp-table}
\end{table*}

\section{Problem Statement} \label{sec:problem statement}
\paragraph{Challenge.} 
The primary obstacle in enhancing the reasoning capabilities of LLMs stems from the \textit{Composition-Knowledge Dichotomy}. This dichotomy has led to a marked polarization in existing research: composition-based approaches (e.g., least-to-most, plan-and-solve, etc.) prioritize well-organized reasoning structures to enhance \textit{compositionality}, whereas knowledge-based approaches (e.g., analogical prompting, self-knowledge explicitation, etc.) focus on grounded evidence to enhance \textit{knowledgeability}. Such polarization makes LLMs vulnerable to either exquisite hallucination or erratic deduction. Specifically, composition-based approaches risk affirming fallacies due to insufficient evidence, while knowledge-based ones risk negating facts due to logically fragmented reasoning.

\paragraph{Formulation.}
The primary goal of our work is to enhance the reasoning capabilities of LLMs in QA tasks by resolving the composition-knowledge dichotomy. To this end, we focus on concretizing the underlying propositions of the questions.

% we focus on minimizing two types of errors: Type I error (affirming a fallacy) and Type II error (negating a fact).

Specifically, we formulate a proposition taxonomy that systematically aligns the \textit{logical mode} of statement (i.e., \textit{compositionality}) with the \textit{truth value} of statement (i.e., \textit{knowledgeability}). Here, the logical mode defines a proposition's syntax as either affirmative ($+$) or negative ($-$), while truth value determines whether the proposition is a fact ($1$) or a fallacy ($0$). Based on this taxonomy, we define proposition categories as follows:
% \begin{enumerate}[label=(\arabic*), leftmargin=*]
%     \item \textbf{True Positive} $(+, 1)$ \textit{(Affirming a Fact)}: Valid inclusion of true information.
%     \item \textbf{True Negative} $(-, 0)$ \textit{(Negating a Fallacy)}: Valid exclusion of false information.
%     \item \textbf{False Positive} $(+, 0)$ \textit{(Affirming a Fallacy)}: Erroneous inclusion of false information. This corresponds to a Type I error.
%     \item \textbf{False Negative} $(-, 1)$ \textit{(Negating a Fact)}: Erroneous exclusion of true information. This corresponds to a Type II error.
% \end{enumerate}
\begin{enumerate}[label=(\arabic*), leftmargin=*]
    \item \textbf{True Positive} $(+, 1)$ \textit{(Affirming a Fact)}: Valid inclusion of true information.
    \item \textbf{True Negative} $(-, 0)$ \textit{(Negating a Fallacy)}: Valid exclusion of false information.
    \item \textbf{False Positive} $(+, 0)$ \textit{(Affirming a Fallacy)}: Erroneous inclusion of false information.
    \item \textbf{False Negative} $(-, 1)$ \textit{(Negating a Fact)}: Erroneous exclusion of true information.
\end{enumerate}
\noindent Table \ref{tab1: cpp-table} provides the definitions and examples of the concretized propositions by category. For further details, please refer to Appendix \ref{apdx: details-of-definitions}.

\section{Method} \label{sec: method}

\begin{figure*}[h!]
    \centering
    \includegraphics[width=\textwidth]{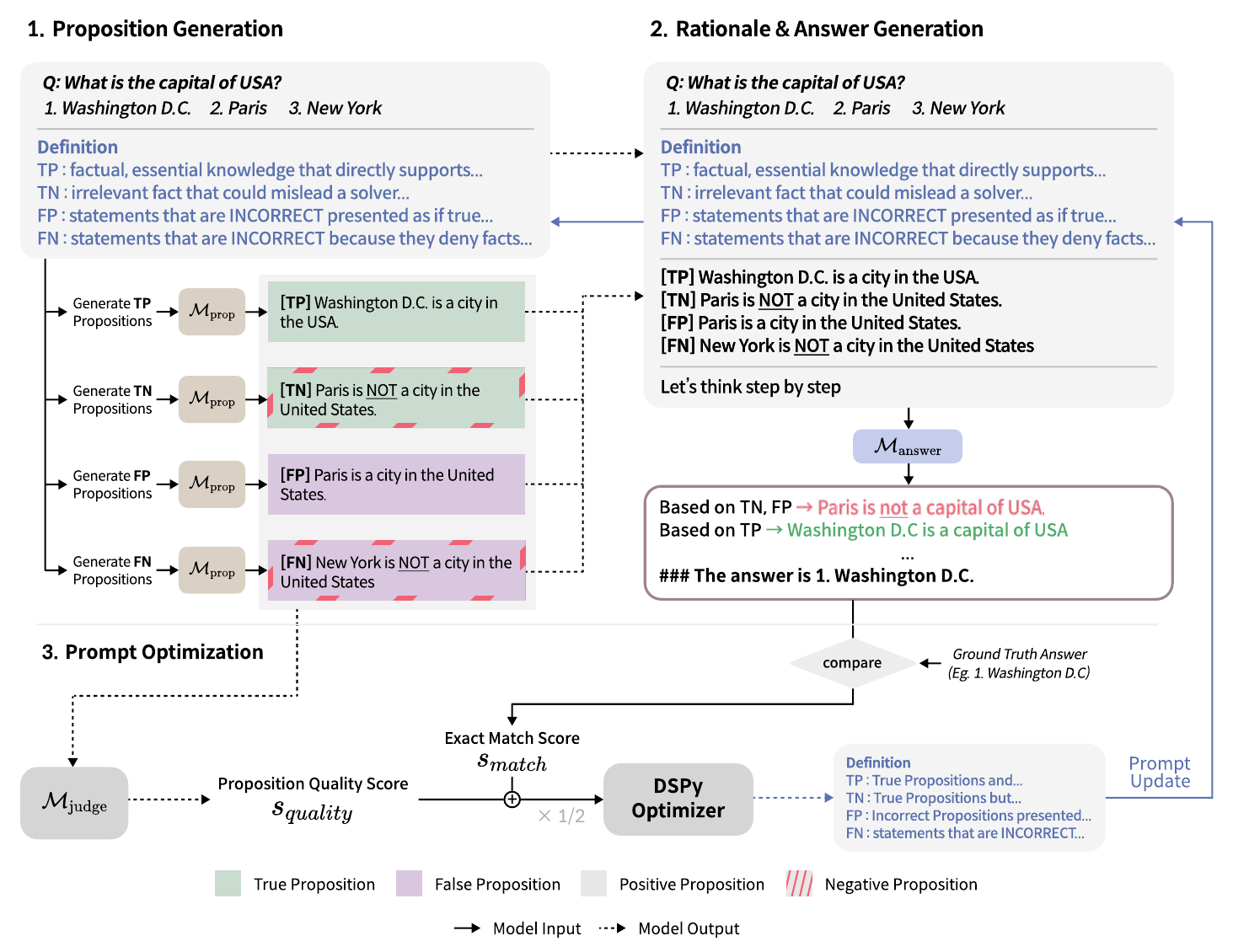}
    \caption{\textbf{Overview of the Concretized Proposition Prompting (CPP) Framework.} The CPP framework consists of three stages: (1) Proposition Generation, where relevant propositions are concretized into four categories (TP, TN, FP, FN) to uncover latent reasoning paths; (2) Rationale \& Answer Generation, where the generated propositions guide the model to derive the final answer; and (3) Prompt Optimization, which uses the DSPy optimizer to jointly update prompts based on a total reward of proposition quality and answer accuracy.}
    \label{fig:figure_1}
\end{figure*}

In this section, we present \textbf{Concretized Proposition Prompting (CPP)}, a simple yet powerful prompting technique designed to enhance the reasoning capability of LLMs. This method systematically concretizes propositions relevant to a question based on proposition categories. Figure \ref{fig:figure_1} describes three stages of the CPP framework.

In the first stage, the proposition model, $\mathcal{M}_{\text{prop}}$, is tasked with generating concrete propositions, $\hat{p} \sim \mathcal{M}_{\text{prop}}(\phi_{\text{prop}}, q)$, by taking the question $q$ and the proposition prompt, $\phi_{\text{prop}}=\left[ \mathcal{I}_{\text{task}}^{(\text{prop})}, \mathcal{I}_{\text{def}}^{(\text{prop})} \right]$, where $\mathcal{I}_{\text{task}}^{(\text{prop})}$ and $\mathcal{I}_{\text{def}}^{(\text{prop})}$ specify task description and the definitions of propositions, respectively. For each proposition category $c$, up to $M$ propositions are generated to build a proposition set $\mathcal{P}$:
\begin{align*}
    \mathcal{P} = \bigcup_{c \in \mathcal{C}} \left\{ \hat{p}_{m}^{(c)} \right\}_{m=1}^{M_c}
\end{align*}
where $\mathcal{C} = \{\text{TP, TN, FP, FN}\}$, and $M_c$ represents the number of propositions for category $c$.\footnote{$\mathcal{M}_{c}$ was set to 5 for all $c$.} To build the proposition set, we design a category-specific prompt $\phi_{\text{prop}}^{c}$ and query the proposition model separately.\footnote{In our pilot study, the use of category-specific prompts tends to generate correct propositions more often than a category-agnostic prompt. See Appendix \ref{apdx: category-agnostic prompt vs. category-specific prompt} for details.} Figure \ref{fig:optimized-proposition-prompt} provides the details.

In the second stage, the proposition set $\mathcal{P}$ is delivered to the answer model $\mathcal{M}_{\text{ans}}$ along with the question $q$ and an answer prompt $\phi_{\text{ans}}=\left[ \mathcal{I}_{\text{task}}^{(\text{ans})}, \mathcal{I}_{\text{cot}}^{(\text{ans})} \right]$, where $\mathcal{I}_{\text{task}}^{(\text{ans})}$ and $\mathcal{I}_{\text{cot}}^{(\text{ans})}$ are the task and CoT instructions, respectively.\footnote{Figure \ref{fig:optimized-answer-prompt} presents the details of the answer prompt.} The task instruction describes the task of the answer model, while the CoT instruction specifies the CoT trigger: \textit{``Let's think step by step.''} The generated outcome from the model is then given by:
\begin{align*}
    \hat{y}_{1:T} = \big[ \underbrace{\hat{r}^{(1)}, \cdots, \hat{r}^{(K)}}_{\ = \ \hat{y}_{1:t-1}}, \hat{y}_{t:T} \big] \sim \mathcal{M}_{\text{ans}}(\phi_{\text{ans}}, q, \mathcal{P})
\end{align*}
where $\left\{\hat{r}^{(k)} \right\}_{k=1}^{K}$ is the chain of thoughts referred to as rationales, and $\hat{y}_{1:T}$ denotes the entire response consisting of the reasoning, $\hat{y}_{1:t-1}$, and conclusion, $\hat{y}_{t:T}$. Note that the model determines reasoning length $K$ by itself. The answer $\hat{a}$ is then parsed from the entire response, $\hat{a}=\operatorname{extract}(\hat{y}_{1:T})$, where the parsing function $\operatorname{extract}(\cdot)$ is implemented by regular expressions.

The third stage focuses on optimizing the proposition prompt $\phi_{\text{prop}}$ and answer prompts $\phi_{\text{ans}}$, based on the DSPy framework. The optimization process aims to maximize 1) the quality of the concretized propositions $\mathcal{P}$ and 2) the accuracy of final answers $\hat{a}$. To measure the quality of propositions, we introduce the judge model $\mathcal{M}_{\text{judge}}$, essentially equivalent to the LLM-as-a-Judge \citep{zheng2023judging}. The judge model predicts numeric scores based on the rubric specified in the judge prompt $\phi_{\text{judge}}$:\footnote{The details of the prompt are presented in Figure \ref{fig:judge-prompt}.}
\begin{align*}
    s_{\text{quality}} = \mathcal{M}_{\text{judge}}(\phi_{\text{judge}}, q, \mathcal{P}) \ .
\end{align*}
Here, $s_{\text{quality}} \in [0, 1]$ indicates the quality score of the concretized propositions. For answer accuracy, we compute the exact match score, formally represented by the indicator function:
\begin{align*}
    s_{\text{match}} = \mathbb{I}[\hat{a}=a] =
    \begin{cases}
        1 & \text{if } \hat{a} = a \\
        0 & \text{otherwise}
    \end{cases}
\end{align*}
where $a$ and $\hat{a}$ are the correct and predicted answers. The two scores---$s_{\text{quality}}$ and $s_{\text{match}}$---represent compositionality and knowledgeability, respectively. By integrating them, we define a total reward function $\mathcal{R}_{\text{total}}$ that equally weighs both axes of compositionality and knowledgeability:
\begin{align*}
    \mathcal{R}_{\text{total}}(\hat{a}, \mathcal{P}, a) \triangleq \frac{s_{\text{quality}} + s_{\text{match}}}{2} \ ,
\end{align*}
which prevents the proposition model from generating hallucinated propositions from which the answer model happens to result in correct answers. 

Building on Section \ref{sec:dspy_optimization}, we jointly optimize the first and second stages w.r.t. the unified prompt parameters $\phi = [\phi_{\text{prop}}; \phi_{\text{ans}}]$. Formally, the optimization objective is defined to identify the optimal parameters $\phi^{*}$ that maximize the expected value of $\mathcal{R}_{\text{total}}$ over the data distribution, $(q, a)\sim\mathcal{D}$, and the model trajectory, $(\mathcal{P}, \hat{y})\sim\mathcal{M}_{\phi}(\cdot\mid q)$:
\begin{align*}
    \phi^* = \operatorname*{argmax}_{\phi} \mathbb{E}_{\substack{(q,a)\sim\mathcal{D} \\ (\mathcal{P}, \hat{y})\sim\mathcal{M}_{\phi}(\cdot\mid q)}}
    \left[ \mathcal{R}_{\text{total}}(\hat{a}, \mathcal{P}, a) \right],
\end{align*}
where the model trajectory includes the set of concretized propositions $\mathcal{P}$ and rationalized response $\hat{y}$, from which the final answer is derived as $\hat{a} = \operatorname{extract}(\hat{y})$. Through optimization, CPP balances compositionality with knowledgeability. Figure \ref{fig:figure_1} describes a part of the prompt being optimized. The fully optimized prompts are described in Figures \ref{fig:optimized-proposition-prompt} and \ref{fig:optimized-answer-prompt}.

\section{Experiments}

\begin{figure*}[h]
\centering
\includegraphics[width=\textwidth]{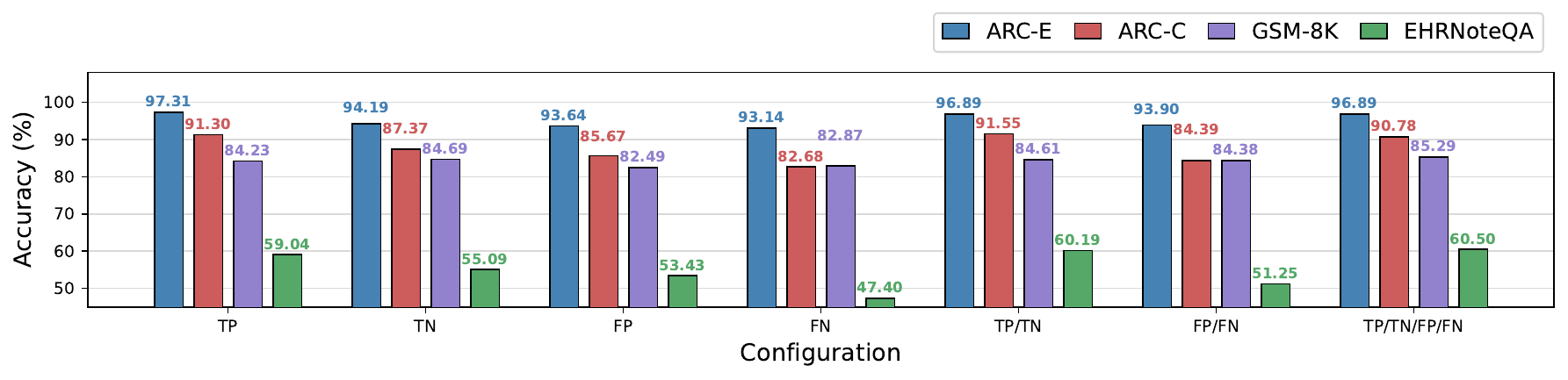}
\caption{\textbf{Accuracy Distribution across  Proposition Category Configuration.} The figure illustrates the accuracy distribution for each configuration, measured by exact match scores across four benchmark datasets.}
\label{fig2: accuracy_distribution}
\end{figure*}

\begin{figure}[t]
    \centering
    \begin{subfigure}[t]{0.48\columnwidth}
        \centering
        \includegraphics[width=\linewidth]{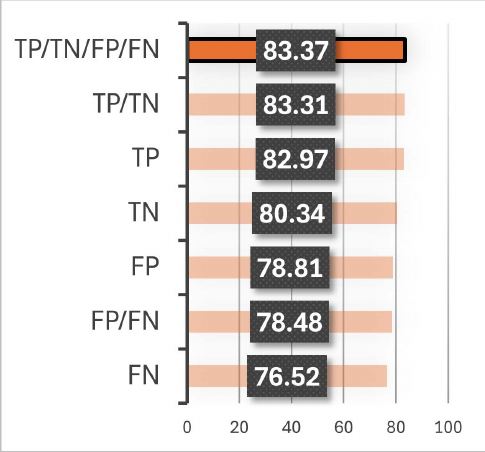}
    \end{subfigure}
    \hfill
    \begin{subfigure}[t]{0.48\columnwidth}
        \centering
        \includegraphics[width=\linewidth]{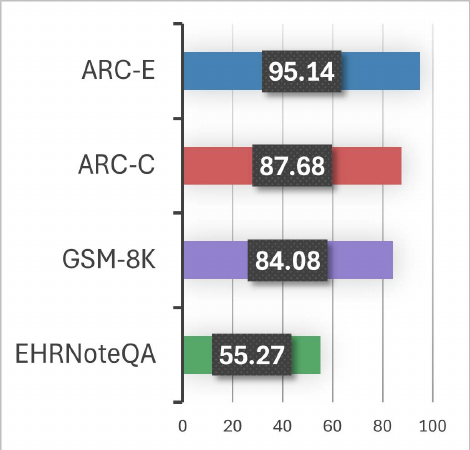}
    \end{subfigure}
    \caption{\textbf{Avg. Accuracy per Category and Dataset.}}
\label{fig3: avg_accuracy_bar}
\end{figure}

\paragraph{Types of Experiments.}
We conducted three distinct sets of experiments: (i) \textbf{ablation study} to explore the optimal configuration of proposition categories; (ii) \textbf{comparison study} to validate the superiority of the CPP; and (iii) \textbf{scalability study} to analyze the performance changes across various foundation models and parameter sizes.

\paragraph{Datasets.}
We evaluate CPP on eight datasets spanning three domains (commonsense, math, medicine). First, we assess general reasoning capabilities using commonsense reasoning benchmarks---ARC-E, ARC-C \citep{Clark2018ThinkYH}, MMLU-Pro \citep{NEURIPS2024_ad236edc}, and CSQA \citep{talmor-etal-2019-commonsenseqa}---which require the coordination of general knowledgeability and basic compositionality. To rigorously evaluate the compositionality, we experiment with GSM-8K \citep{cobbe2021} and MATH \citep{hendrycks2021measuring} datasets, as they demand organized logical deductions. Finally, to assess the knowledgeability, we employ medical benchmarks including EHRNoteQA \citep{NEURIPS2024_e15c4aff} and MedXpertQA \citep{zuo2025medxpertqa}, where factual correctness is prioritized. The details of each dataset are provided in Appendix \ref{apdx: details-of-benchmark-datasets}.

% \subsection{Datasets}
% We evaluate the proposed framework on eight datasets spanning three distinct domains to verify its versatility across varying levels of reasoning complexity. 
% We initially anchor our evaluation in commonsense reasoning to test the model's grasp of general world knowledge and scientific principles. 
% This category includes ARC-C and ARC-E~\citep{Clark2018ThinkYH}, which distinguish between complex reasoning and basic retrieval, alongside MMLU-Pro and CommonsenseQA for a robust assessment of broad semantic understanding. 
% Moving beyond general knowledge to require rigorous logical structures, we incorporate mathematical reasoning tasks via GSM-8K~\citep{cobbe2021} and MATH. 
% Finally, to assess the framework's applicability in high-stakes professional environments where precise knowledgeability is paramount, we extend our validation to the medical domain using EHRNoteQA~\citep{NEURIPS2024_e15c4aff} and MedXpertQA. 

\paragraph{Models.}
We employ a total of seven models in our experiments. In the ablation study, we experiment with \texttt{Llama-3.1-8B-Instruct} \citep{grattafiori2024llama} to search for the best configuration of proposition category. All methods used in the comparison study are also based on \texttt{Llama-3.1-8B-Instruct}. For the scalability study, we employ a variety of foundation models, such as \texttt{Qwen2.5-7B-Instruct} \citep{yang2024qwen2.5}, \texttt{Qwen2-7B-Instruct} \citep{yang2024qwen2}, \texttt{Gemma-3-12b-it} \citep{team2025gemma}, \texttt{Phi-3-small-8k-instruct} \citep{abdin2024phi}, \texttt{Mistral-3-8B-Instruct} \citep{jiang2023mistral7b}.

% - Llama-3.1-8B-Instruct
% - Qwen2-7B-Instruct
% - Qwen2.5-7B-Instruct
% - Mistral-3-8B-Instruct
% - Gemma-3-12b-it
% - Phi-3-small-8k-instruct (7B)

\paragraph{Baselines.}
We primarily compare the CPP with zero-shot CoT \citep{kojima2022, wei2022} and direct answer baselines. This comparison enables us to analyze the relative superiority of the CPP to the standard approaches; it demonstrates how effectively concrete propositions enhance the reasoning capabilities of LLMs. Additionally, we include few-shot prompting \citep{brown2020} as a baseline. This inclusion aims to verify that concretizing propositions specific to the target question provides better guidance for LLM reasoning than relying on a few arbitrary question-answer pairs.

\paragraph{Evaluation.}
To compare the baselines, other advanced prompting methods, and CPP in test-time performance, we evaluate the exact match score for all datasets as in previous studies. The score is obtained according to the parsing mechanism with a predefined regular expression. The only exception is the MATH dataset. Since the answers in MATH often involve symbolic or irrational solutions, simply parsing strings with a regular expression cannot account for mathematical equivalence.\footnote{Different strings can represent the same mathematical identity, such as symbolic permutations ($x+y$ vs. $y+x$) or the relationship between exact values and their decimal approximations ($\sqrt{2}$ vs. $1.4142...$).} To bypass this issue, we use another LLM-as-a-Judge instead.

% To address these issues, we utilize the \texttt{Math-Verify} library developed by Hugging Face,\footnote{https://github.com/huggingface/Math-Verify} which ensures a rigorous assessment of mathematical equivalence.

\paragraph{Implementation.}
We employ the DSPy framework with the Ollama engine during optimization, while vLLM \citep{kwon2023efficient}, a high-throughput inference package, is used during evaluation. For $\mathcal{M}_{\text{prop}}$ and $\mathcal{M}_{\text{ans}}$, we use \texttt{Qwen2.5-72B-Instruct} and \texttt{Llama-3.1-8B-Instruct}, respectively, and set up greedy decoding so that proposition and answer generation rely solely on prompt updates. In contrast, the DSPy optimizer, responsible for updating prompts, and the judge model $\mathcal{M}_{\text{judge}}$ utilize stochastic decoding ($T=1.0$) to explore diverse prompts for efficient optimization and to perform extensive reasoning for accurate judgments, respectively. Note that the DSPy optimizer and the judge model are implemented by \texttt{gpt-oss-120B}.

% Lastly, the proposition model and answer model of CPP are implemented with \texttt{Qwen2.5-72B-Instruct} and \texttt{Llama-3.1-8B-Instruct}, unless otherwise specified, to consider fair comparisons with baselines or other prompting methods.

\begin{table*}[t]
\small
\centering
\scriptsize                    % small보다 한 단계 더 줄이면 덜 깨짐
\setlength{\tabcolsep}{4pt}
\renewcommand{\arraystretch}{1.15}
\resizebox{\textwidth}{!}{%
\begin{tabular}{lcccccccc}
\toprule
Method
& \multicolumn{4}{c}{Commonsense}
& \multicolumn{2}{c}{Math}
& \multicolumn{2}{c}{Medicine} \\
\cmidrule(lr){2-5}\cmidrule(lr){6-7}\cmidrule(lr){8-9}
& ARC-E & ARC-C & MMLU-Pro & CSQA & GSM-8K & MATH & EHRNoteQA & MedXpertQA \\
\midrule
Zero-shot Direct Answer \citep{brown2020}   & 92.5$^{\dagger}$ & 82.6$^{\dagger}$ & 38.0$^{\dagger}$ & 74.9$^{\dagger}$ & 18.5$^{\dagger}$ & 14.2 & \underline{60.2} & 14.2 \\
Zero-shot CoT \citep{wei2022}  & 93.0$^{\dagger}$ & 86.0$^{\dagger}$ & 44.8$^{\dagger}$ & 68.5$^{\dagger}$ & \underline{85.4}$^{\dagger}$ & 48.4 & 56.5 & 13.8 \\
Few-shot Direct Answer \citep{brown2020} & 91.0 & 78.1 & 38.0$^{\dagger}$ & 75.2 & 20.1$^{\dagger}$ & 16.2 & - & - \\
\midrule
Plan-and-Solve \citep{wang-etal-2023-plan} &  94.8  & 87.0  & 47.5 & \textbf{77.8} & \textbf{85.5} & 46.4 & 56.4 & 14.5 \\
Rephrase-and-Respond \citep{deng2023rephrase} & 95.0  & 86.6  & 47.7 & 76.2 & 71.5 & 32.6 & 56.5 & 13.9 \\
System-2-Attention \citep{weston20232attentionisneed} & 93.6  & 86.3  & 43.4 & 74.3 & 61.7 & 40.2 & 52.8 & \underline{14.8} \\
Thread-of-Thought \citep{zhou2023threadthoughtunravelingchaotic}  & \underline{95.4} & \underline{87.5}  & \underline{48.8} &  \underline{77.6} & 85.2 & \underline{50.0} & 57.3 & 14.2 \\
Analogical \citep{yasunaga2023} & 87.0  & 78.2  & 41.4 & 68.7 & 73.5 & 40.8 & 54.7 & 12.4 \\
Re-reading \citep{xu-etal-2024-reading} & 94.2  & 87.5  & 46.3 & 74.9 & 82.8 & 49.2 & 57.6 & 13.1 \\
SKE-Learn \citep{huang2025} & 92.0  & 82.6  & 39.1 & 75.4 & 75.8 & 41.6 & 50.9 & 14.2 \\
Least-to-Most \citep{zhou2022} & 91.9 & 82.8 & 38.1 & 72.8 & 49.1 & 30.8 & 49.2 & 13.1 \\
Meta-Prompting \citep{zhang2023meta} & - & - & - & - & 85.2 & 49.6 & - & - \\

\midrule

CPP (with TP/TN/FP/FN) & \textbf{96.9} & \textbf{90.8} & \textbf{51.9} & 77.2 & 85.3 & \textbf{50.4} & \textbf{60.5} & \textbf{15.5} \\
% CPP (Qwen2-7B-Instruct)      & \textbf{91.5} & \textbf{83.7} & \textbf{41.2} & \textbf{73.1} & \textbf{-} & \textbf{-} & \textbf{48.5} & \textbf{11.2} \\

\bottomrule
\end{tabular}%
}
\caption{\textbf{Overall Results of Comparative Evaluation.} The numbers represent accuracy measured by exact match scores for all datasets, excluding MATH. The top and second-best performance for each dataset are marked in \textbf{bold} and \underline{underlined}, respectively. Results with $^{\dagger}$ symbol are retrieved from \citet{sprague2025to}, while all the other numbers are from our own implementation.}

\label{tab2: comparison-table}
\end{table*}

\section{Results}

\begin{figure*}
\includegraphics[width=\linewidth]{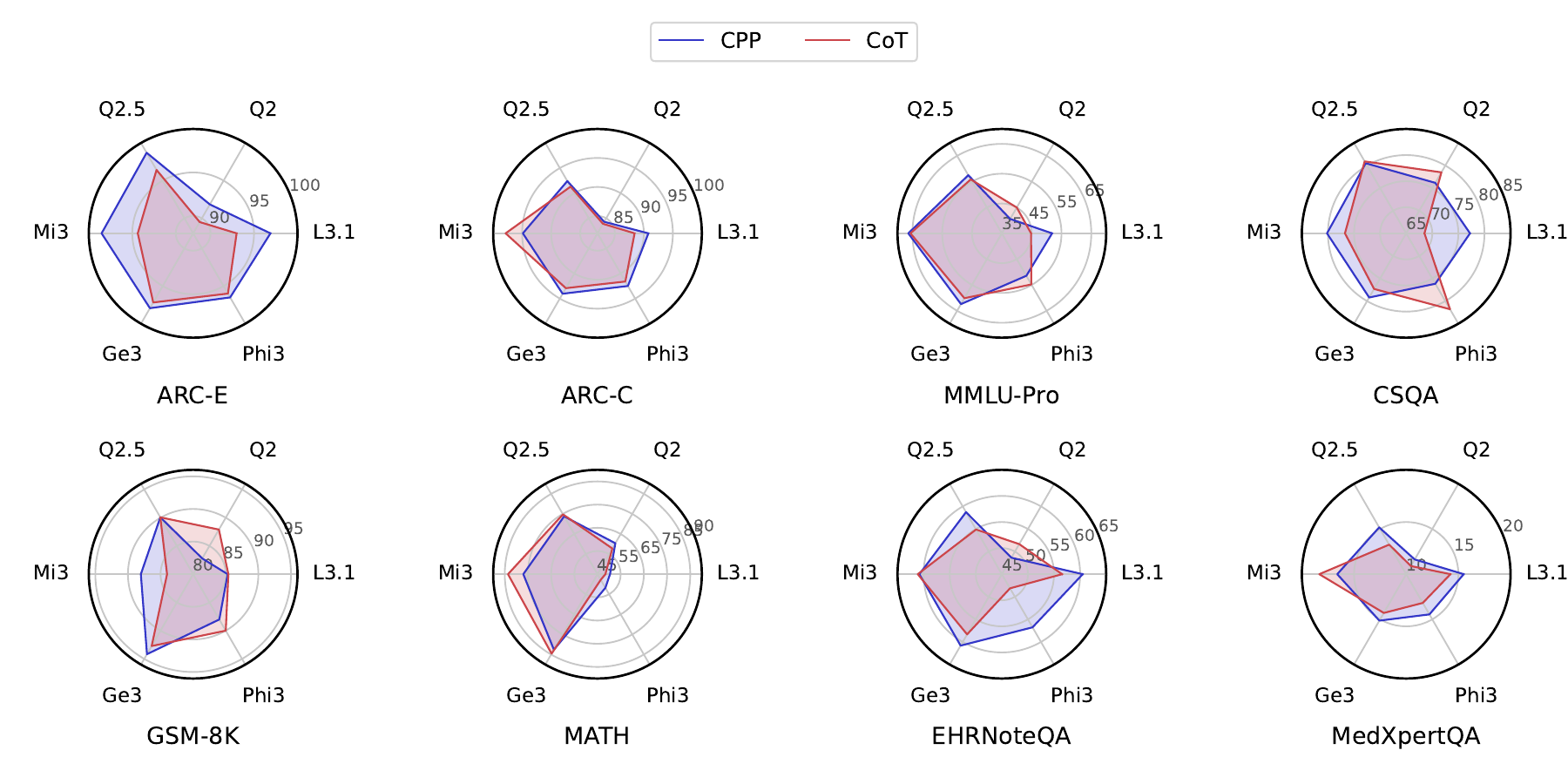}
\caption{\textbf{Performance Analysis across Foundation Models.} The figure illustrates the performance variability across different FMs. Each vertex of radar chart corresponds to a specific model; by clockwise, Q2.5, Q2, L3.1, Phi3, Ge3, and Mi3 denote Qwen2.5-7B, Qwen2-7B, Llama-3.1-8B, Phi-3-7B, Gemma-3-12B, and Mistral-3-8B, respectively. The blue and red lines represent the performance of CPP and Zero-shot CoT, respectively.}
\label{fig4: foundation-model-analysis}
\end{figure*}

% 2. 같은 Model Famaily내 7B, 14B, ..., 70B 별로 실험을 진행
% - 선 그래프 Figure 로 보여주기(Chain-of-Thought Prompting Elicits Reasoning in Large Language Models 논문 참고)

% - Llama-3.1-8B-Instruct
% - Llama-3.1-70B-Instruct

% - Qwen2-7B-Instruct
% - Qwen2-72B-Instruct

% ---------------------------------------
% - Qwen2.5-7B-Instruct
% - Qwen2.5-14B-Instruct
% - Qwen2.5-32B-Instruct
% - Qwen2.5-72B-Instruct
% ---------------------------------------

% - Mistral-3-3B-Instruct
% - Mistral-3-8B-Instruct
% - Mistral-3-14B-Instruct

% - Gemma-2-2b-it
% - Gemma-3-12b-it
% - Gemma-2-27b-it

% - Phi-3-small-4k-instruct (3.8B)
% - Phi-3-small-8k-instruct (7B)
% - Phi-3-small-128k-instruct (7B)
% - Phi-3-medium-4k-instruct (14B)
% - Phi-3-medium-128k-instruct (14B)
\begin{figure*}[t]
\centering
\includegraphics[width=\linewidth]{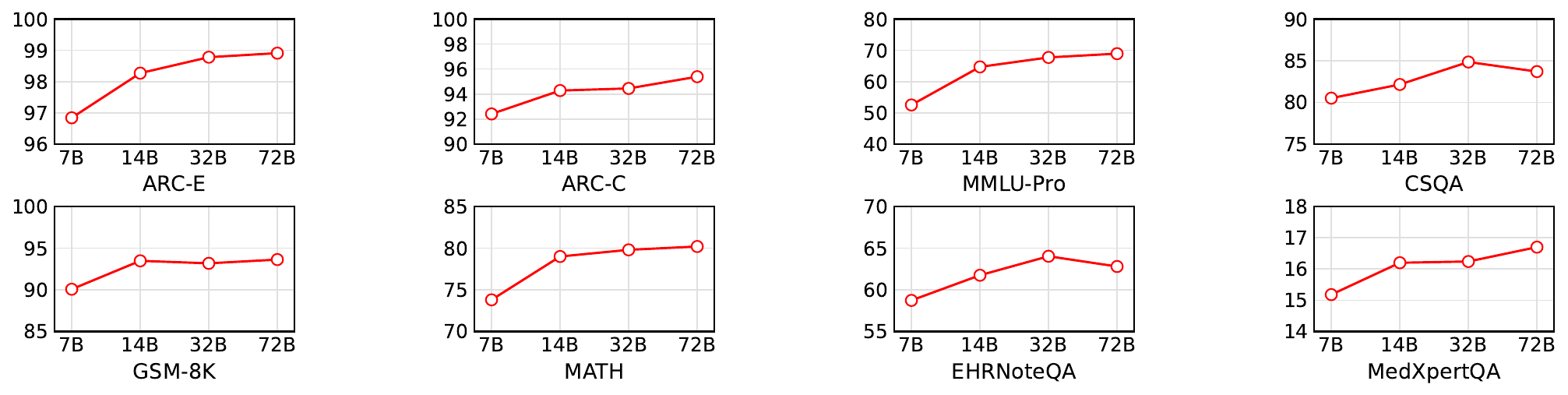}
\caption{\textbf{Performance Analysis across Parameter Sizes.} The line graphs illustrate the changes in performance of Qwen2.5 as parameter sizes increase from 7B to 72B.}
\label{fig5: parameter-size-analysis}
\end{figure*}

\paragraph{Ablation Study.}
The core idea of CPP is to concretize question-relevant propositions. To identify the most effective proposition setup, we validate a variety of proposition configurations. 

Figure \ref{fig2: accuracy_distribution} illustrates the accuracy distribution for each configuration, measured by exact match scores. The average accuracy per configuration across four benchmark datasets is: 82.97 (TP), 80.34 (TN), 78.81 (FP), 76.52 (FN), 83.31 (TP/TN), 78.48 (FP/FN), and 83.37 (TP/TN/FP/FN), which is shown in the left-sided plot of Figure \ref{fig3: avg_accuracy_bar}. The results indicate that the $\text{TP/TN/FP/FN}$ configuration is optimal, as it achieves the highest average accuracy across all benchmark datasets. Accordingly, all subsequent experiments in this study are conducted using the TP/TN/FP/FN configuration. 

In addition, the right-hand plot of Figure \ref{fig3: avg_accuracy_bar} shows that CPP performs better in the order of the commonsense, math, and medicine domains, where the average accuracy per dataset is given by 95.14 (ARC-E), 87.68 (ARC-C), 84.08 (GSM-8K), and 55.27 (EHRNoteQA). This may be due to the disparity in the explicitness of logical structures and the depth of required knowledge across these domains; while commonsense and mathematical reasoning rely on relatively transparent propositions, the medical reasoning requires handling clinical records where the content is often ambiguous or context-dependent (e.g., uncertain causalities or unestablished clinical facts).

% 1. TP/TN/FP/FN 어떤 조합을 프롬프트에 고려할지에 따라 성능이 어떻게 변화하는지 분석
% - ARC-C, ARC-E 로만 평가
% - TP, TN, FP, FN, ..., bar graph Figure 로 보여주기
% 결과: TP/TN/FP/FN/... 조합들 중에 configuration 결정

% 2. (TP/TN/FP/FN을 서로 다른 LLM으로 뽑는 경우 vs 하나의 LLM이 모든 명제를 한번에 뽑는 경우)의 바교 : ARC-C와 ARC-E만으로 실험
% - ARC-C, ARC-E 로만 평가
% - LLM-as-a-Judge(명제의 일관성 평가, GPT-OSS 120B) + QA Score bar graph Figure 로 보여주기
% 결과: single-LLM vs. multi-LLM 중 configuration 결정
% \begin{table}[t]
% \small
% \centering
% \setlength{\tabcolsep}{3pt}
% \renewcommand{\arraystretch}{1.15}
% \begin{tabularx}{\columnwidth}{@{}Ycccc@{}}
% \toprule
% \multirow{2}{*}{Setting} & \multicolumn{2}{c}{ARC-C} & \multicolumn{2}{c}{ARC-E} \\
% \cmidrule(lr){2-3}\cmidrule(lr){4-5}
% & Judge$\uparrow$ & Answer (\%)$\uparrow$ & Judge$\uparrow$ & Answer (\%)$\uparrow$ \\
% \midrule
% Single-LLM & 0.76 & 83.1 & 0.78 & 89.1 \\
% Multi-LLM & \textbf{0.88} & \textbf{84.9} & \textbf{0.90} & \textbf{90.2} \\
% \bottomrule
% \end{tabularx}
% \caption{Ablation study 2}

% \end{table}

\paragraph{Comparison Study.}
% 1. 위 분석을 통해 확정한 configuration을 갖고 dataset들에 대해서 baseline들과 비교
% - baselines : Direct Answer vs. Few-shot prompting vs. zero-shot CoT vs. CPP 
% - Table 로 보여주기
% - 예시 Table도 보여주기 (어떻게 다르게 추론하고 정답에 도달하는지 예시를 보여주는 것)

To validate the superiority of CPP against various baselines and advanced prompting methods, we conducted a comparative evaluation where all methods were initialized with \texttt{Llama-3.1-8B-Instruct}. For CPP, we employed category-specific prompts with TP/TN/FP/FN configuration. Table \ref{tab2: comparison-table} shows that CPP outperforms comparison methods across most datasets. We summarize the comparison models in Table \ref{tab4: classification-of-comparison-methods}.

In commonsense benchmarks, which demand a balance between composition- and knowledge-based approaches, CPP achieves top accuracy on three out of four benchmarks---ARC-E (96.9\%), ARC-C (90.8\%), and MMLU-Pro (51.9\%)---suggesting that CPP effectively balances compositionality with knowledgeability. In math benchmarks, where deductive reasoning is prioritized, CPP is competitive on GSM-8K (85.3\%) and superior on MATH (50.4\%), running far ahead of zero-shot and few-shot direct answer models. These results imply that CPP enhances the reasoning capabilities of LLMs along the compositionality axis, proving concrete propositions more effective than a few arbitrary examples. Tables \ref{tab4: arc_e_full_text}--\ref{tab7: math_full_text (cont.)} compare the generated examples by CoT and CPP on ARC-E, ARC-C, GSM-8K, and MATH datasets. 

The performance of CPP is also pronounced in medical benchmarks where precise knowledge is paramount. On EHRNoteQA, CPP achieves the best result (60.5\%). Notably, Zero-shot CoT (56.5\%) performs worse than Zero-shot Direct Answer (60.2\%), which is likely due to hallucinated CoT. These results suggest that CPP equips LLMs with anti-hallucination behavior and preserves their reasoning capabilities along the knowledgeability axis, keeping LLMs from drifting into hallucinated clinical narratives. Table \ref{tab8: ehrnoteqa_full_text_521} provides the generated examples in the EHRNoteQA dataset. 

% To sum up, CPP successfully minimizes Type II and Type I errors by utilizing concrete propositions. Specifically, the remarkable performance in math benchmarks suggests that CPP enhances the reasoning capabilities of LLMs along the compositionality axis by preventing Type II errors (negating facts, FN) to reduce logically fragmented reasoning. Similarly, the superior results in medical benchmarks imply that CPP strengthens the knowledgeability axis by preventing Type I errors (affirming fallacies, FP) to suppress hallucination. As a result, CPP resolves the composition-knowledge dichotomy and ensures reasoning that is both logically organized and factually grounded.

To sum up, CPP successfully resolves the composition-knowledge dichotomy by utilizing concrete propositions. Specifically, the remarkable outcomes in the commonsense benchmark prove that CPP balances compositionality with knowledgeability. The competitive performance in math suggests that CPP enhances reasoning along the compositionality axis; similarly, superior results in medical benchmarks imply that it strengthens knowledgeability. Collectively, these findings position CPP as a fundamental paradigm for achieving reasoning that is both logically organized and factually grounded.

\paragraph{Scalability Study.}
Lastly, we analyzed how the performance of CPP scales across different foundation models and parameter sizes compared to the standard baseline, Zero-shot CoT.

First, we investigated the model scalability of CPP by evaluating its performance across six foundation models, including Qwen, Llama, Phi, Gemma, and Mistral. Figure \ref{fig4: foundation-model-analysis} reveals that the performance of CPP changes according to the models and datasets used. Building answer models upon Qwen2.5, Llama-3.1, Gemma-3, the CPP outperforms or almost matches CoT on all benchmark datasets, otherwise, CPP and CoT compete against each other depending on the dataset. That is, CPP outperforms in newer models with higher reasoning capacities consistently, which suggests better scalability to models than CoT.

Next, we examined the size scalability of CPP across different parameter sizes of the answer model using the Qwen2.5 suite (7B--72B).\footnote{We selected Qwen2.5 because it performed consistently well in the model scalability experiments and offers a wide variety of sizes for systematic analysis.}  Figure \ref{fig5: parameter-size-analysis} demonstrates that, on most datasets, performance improves near-linearly as parameter size increases. Exceptionally, CSQA and EHRNoteQA reach their peak performance at the 32B scale. We suspect this can happen when larger (answer) models follow wrong instructions so strictly that it reflects incorrect propositions in their final answers. This implies that larger size does not always guarantee better performance, emphasizing the need for further research on the proposition model to improve the quality of proposition generation.

\section{Conclusion}
In this study, we presented Concretized Proposition Prompting (CPP) to enhance the reasoning capabilities of LLMs. By explicitly concretizing relevant propositions, CPP has achieved the highest performance on six out of eight benchmark datasets, shown scalable to various foundation models, and with increasing parameter sizes. These results demonstrate that CPP successfully balances compositionality with knowledgeability, regardless of models and parameter sizes, being a fundamental paradigm that bridges the gap between composition-based and knowledge-based approaches. Consequently, CPP resolves the composition-knowledge dichotomy by providing a solid foundation for logically organized and factually grounded reasoning.

% and provides a solid foundation for logically organized and factually grounded reasoning.

% Furthermore, while performance generally scales near-linearly with parameter size,

% though its explicit structure can sometimes degrade its performance in complex mathematical reasoning. Nonetheless, CPP still serves as a fundamental paradigm that bridges the gap between composition-based and knowledge-based approaches, providing a solid foundation for logically organized and factually grounded reasoning.

\section{Limitations}
% Despite its strengths, CPP has several limitations that provide opportunities for future research. First, in complex, multi-step problems like the MATH dataset, the explicit structure of category-specific prompts can hinder the organic integration of propositions. Second, its success is not yet universal; the performance depends on the specific foundation model. Third, we found that very large models (72B) can follow instructions so strictly that they might include even incorrect propositions in their final answers. Fourth, in the medical domain, performance remains relatively lower because clinical records are often ambiguous or context-dependent. In such cases, simply extracting propositions may not be enough to resolve all clinical uncertainties. Finally, the two-step approach---generating propositions first and then the answer---takes more effort than standard single-step methods. Taking these challenges into account, future work should focus on developing a simpler and more flexible framework that maintains strong knowledgeability while improving compositionality across diverse models and complex, ambiguous tasks.

Despite its strengths, CPP has several limitations that provide opportunities for future research. First, its success is not yet universal; the performance changes depending on the models and datasets used. Second, given incorrect propositions, the large answer model may follow the wrong instructions so strictly that it might lead to incorrect propositions in the final answers. Third, in the medical datasets, overall performance remains significantly lower compared to others. This is likely due to the fact that clinical records are often ambiguous or context-dependent (e.g., uncertain causalities or unestablished clinical facts). Finally, the two-step approach---generating propositions first and then the answer---takes more effort than standard single-step methods. Taking these challenges into account, future work should focus on developing a simpler and more flexible framework that balances compositionality with knowledgeability universally applicable to various datasets and models.

% \lipsum[4]

% \section*{Limitations}

% This document does not cover the content requirements for ACL or any
% other specific venue.  Check the author instructions for
% information on
% maximum page lengths, the required ``Limitations'' section,
% and so on.

% \section*{Acknowledgments}

% This document has been adapted
% by Steven Bethard, Ryan Cotterell and Rui Yan
% from the instructions for earlier ACL and NAACL proceedings, including those for
% ACL 2019 by Douwe Kiela and Ivan Vuli\'{c},
% NAACL 2019 by Stephanie Lukin and Alla Roskovskaya,
% ACL 2018 by Shay Cohen, Kevin Gimpel, and Wei Lu,
% NAACL 2018 by Margaret Mitchell and Stephanie Lukin,
% Bib\TeX{} suggestions for (NA)ACL 2017/2018 from Jason Eisner,
% ACL 2017 by Dan Gildea and Min-Yen Kan,
% NAACL 2017 by Margaret Mitchell,
% ACL 2012 by Maggie Li and Michael White,
% ACL 2010 by Jing-Shin Chang and Philipp Koehn,
% ACL 2008 by Johanna D. Moore, Simone Teufel, James Allan, and Sadaoki Furui,
% ACL 2005 by Hwee Tou Ng and Kemal Oflazer,
% ACL 2002 by Eugene Charniak and Dekang Lin,
% and earlier ACL and EACL formats written by several people, including
% John Chen, Henry S. Thompson and Donald Walker.
% Additional elements were taken from the formatting instructions of the \emph{International Joint Conference on Artificial Intelligence} and the \emph{Conference on Computer Vision and Pattern Recognition}.

\bibliography{custom, anthology}

\newpage
\appendix

\section{Details of Definitions} \label{apdx: details-of-definitions}

This section provides the formal grounding for the Concretized Proposition Prompting (CPP) framework. Our objective is to resolve the \textbf{Composition-Knowledge Dichotomy} by aligning the \textbf{Logical Mode} with its \textbf{Veracity}. Let $\mathcal{P}$ be the universe of all possible propositions and $\mathcal{G} \subset \mathcal{P}$ be the set of factually grounded information.

\subsection{The Two Axes of Reasoning}
To resolve the dichotomy, we define independent axes governing how information is structurally organized and factually evaluated:

\begin{definition} \textbf{(Logical Mode: The Axis of Compositionality)}
The Logical Mode $\sigma \in \{+, -\}$ determines the structural syntax of a statement---specifically, whether to include ($+$) or exclude ($-$) information. We define a structural operator $\mathcal{C}$ that transforms a proposition $p$ based on $\sigma$:
\begin{equation}
    \mathcal{C}(p, \sigma) = 
    \begin{cases} 
    p & \text{if } \sigma = + \quad \text{(Affirmation)} \\
    \neg p & \text{if } \sigma = - \quad \text{(Negation)}
    \end{cases}
\end{equation}
\end{definition}

\begin{definition} \textbf{(Veracity: The Axis of Knowledgeability)}
Veracity refers to the factual correctness of $p$, independent of its logical mode. We define a valuation function $v: \mathcal{P} \to \{1, 0\}$ to indicate if a proposition is a fact or a fallacy:
\begin{equation}
    v(p) = 
    \begin{cases} 
    1 & \text{if } p \in \mathcal{G} \quad \text{(Fact)} \\
    0 & \text{if } p \notin \mathcal{G} \quad \text{(Fallacy)}
    \end{cases}
\end{equation}
\end{definition}

\subsection{Taxonomy of Propositions}

By synthesizing the selection of $\sigma$ and the valuation $v(p)$, we define four categories by pairs $(\sigma, v)$. This taxonomy systematically identifies the underlying propositions:

% \begin{itemize}[leftmargin=*]
%     \item \textbf{True Positive (TP)} $(+, 1)$: Successfully affirming a fact. Provides grounded evidence for valid reasoning.
%     \item \textbf{True Negative (TN)} $(-, 0)$: Successfully negating a fallacy. Prevents the model from being misled by misconceptions.
%     \item \textbf{False Positive (FP)} $(+, 0)$: Erroneously affirming a fallacy (\textbf{Type I error}). Leads to hallucinations.
%     \item \textbf{False Negative (FN)} $(-, 1)$: Erroneously negating a fact (\textbf{Type II error}). Leads to logically fragmented reasoning.
% \end{itemize}

\begin{itemize}[leftmargin=*]
    \item \textbf{True Positive (TP)} $(+, 1)$: Successfully affirming a fact. Provides grounded evidence for valid reasoning.
    \item \textbf{True Negative (TN)} $(-, 0)$: Successfully negating a fallacy. Prevents the model from being misled by misconceptions.
    \item \textbf{False Positive (FP)} $(+, 0)$: Erroneously affirming a fallacy.
    \item \textbf{False Negative (FN)} $(-, 1)$: Erroneously negating a fact.
\end{itemize}

% \noindent By including the propositions of all categories---\textbf{TP}, \textbf{TN}, \textbf{FP}, and \textbf{FN}---CPP ensures to integrate compositionality and knowledgeability. Specifically, CPP addresses the composition-knowledge dichotomy by framing it as an error mitigation; eliminating Type I errors involves shifting the model from the erroneous affirmation of fallacies (\textbf{FP}) toward their active rejection (\textbf{TN}). Similarly, mitigating Type II errors involves transforming the negation of essential facts (\textbf{FN}) into their explicit affirmation (\textbf{TP}), preventing critical evidence from being ignored during the deduction process. 

\noindent By including the propositions of all categories---\textbf{TP}, \textbf{TN}, \textbf{FP}, and \textbf{FN}---CPP simultaneously reflects compositionality and knowledgeability. Consequently, this unified control over the four proposition categories allows LLMs to resolve the composition-knowledge dichotomy prevalent in current paradigms.

\section{Category-agnostic vs. Category-specific Prompting} \label{apdx: category-agnostic prompt vs. category-specific prompt}
This section justifies the design choice of $\mathcal{M}_{\text{prop}}$ detailed in Section \ref{sec: method}, the use of separate prompts $\phi_{\text{prop}}^{c}$ for each category $c$. Specifically, we evaluate whether considering category-specific prompts and generating category-specific propositions outperforms a category-agnostic approach that uses a single unified prompt to generate propositions for all categories.

\begin{figure}[!ht]
\centering
\includegraphics[width=\columnwidth]{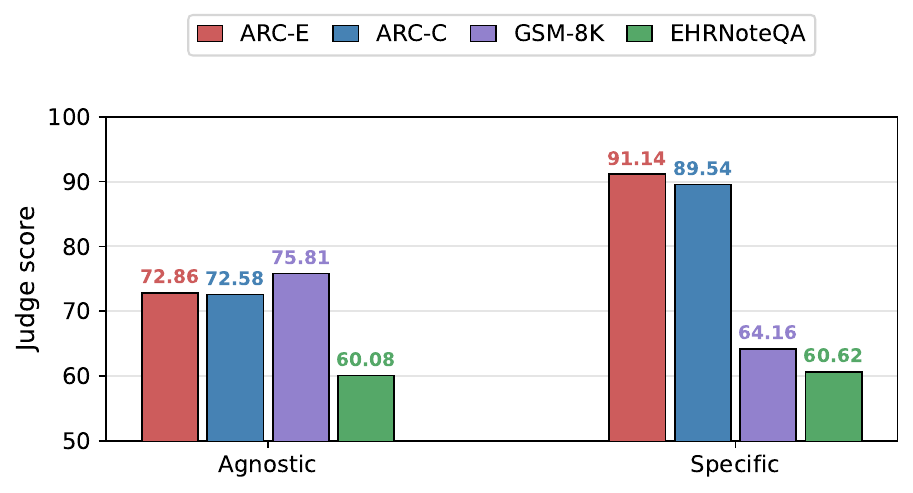}
\caption{Agnostic vs Specific proposition generation. We report Judge score (0-1) on ARC-C/ARC-E, GSM-8K, and EHRNoteQA.}
\label{fig: ablation2_bar}
\end{figure}

Figure \ref{fig: ablation2_bar} compares the judge scores of two distinct prompting approaches---category-specific approach and category-agnostic approach---across the four benchmark datasets. The results indicate that a category-specific prompt significantly outperforms the category-agnostic prompt in the ARC-E and ARC-C datasets. Conversely, on the GSM-8K and EHRNoteQA datasets, the category-specific approach either lagged significantly behind or only slightly outperformed the category-agnostic approach. 

This difference suggests that the best prompting strategy depends on the characteristics of domains. For the commonsense domain (ARC-E and ARC-C), where reasoning relies on relatively independent modular facts, category-specific prompts facilitate a focused extraction of granular details. In contrast, the intricate logic of the math (GSM-8K) and medicine (EHRNoteQA) domains necessitates complex reasoning. We assume that a category-agnostic prompt helps models connect essential information by considering proposition categories altogether. This explains why the category-specific prompt is less effective on GSM-8K and EHRNoteQA compared to ARC-E and ARC-C.

In conclusion, we opted for the category-specific approach despite its lower performance on the GSM-8K dataset. This decision is based on the significant performance gains in ARC-E and ARC-C, which demonstrate that generating propositions for each category separately is more effective for capturing granular and independent facts. While this method may sacrifice performance in GSM-8K, we believe the performance drop is not a fundamental limitation but a challenge that can be overcome through further prompt optimization.

\section{Details of Prompts}
\label{apdx: details-of-prompts}

This section provides a detailed examination of the prompts employed in the CPP framework. Figures \ref{fig:optimized-proposition-prompt} and \ref{fig:optimized-answer-prompt} illustrate the final prompts for the proposition model $\mathcal{M}_{\text{prop}}$ and the answer model $\mathcal{M}_{\text{ans}}$, respectively, which were optimized through the DSPy framework \citep{khattab2023, opsahl-ong-etal-2024-optimizing}.

The optimization process focused on the instructional components (highlighted in gray-blue) to align the models with our specific objectives. For $\mathcal{M}_{\text{prop}}$, the optimization emphasized strictly adhering to the definitions of the four proposition categories (TP, TN, FP, FN) and imposing constraints to prevent information leakage from the answer options. The optimized instructions explicitly direct the model to derive propositions solely from domain principles, independent of the provided options. For $\mathcal{M}_{\text{ans}}$, the optimization refined the task description to ensure that the model effectively integrates the concretized propositions into its reasoning chain before concluding with an answer.

Figure \ref{fig:judge-prompt} presents the judge prompt used to evaluate the quality of generated propositions. Unlike the trainable prompts, this prompt serves as a static reference for the reward function, establishing a rigorous rubric to guide the optimization of the other two models.

% --- Figure 1: Proposition Prompt ---
\begin{figure*}[ht!]
    \centering
    % [Technique] Use adjustbox to scale the entire box down to 90% of text width
    % This keeps aspect ratio but shrinks the size.
    \begin{adjustbox}{width=1.0\textwidth} 
    \begin{promptbox}{Optimized Prompt for $\mathcal{M}_{\text{prop}}$}
    % [Technique] Reduce font size and line spacing inside the box
    \small 
    \renewcommand{\baselinestretch}{0.9}\selectfont 
    
\textcolor{mutedblue}{You are a domain-aware knowledge generator for multiple-choice questions. \\
Given a question, a list of answer options, and a target type (TP, TN, FP, or FN), produce one or more concise, self-contained sentences following these definitions: \\
1. TP (True Positive): factual, essential knowledge that directly supports solving the problem. \\
2. TN (True Negative): factual knowledge that reflects a common misconception or irrelevant fact that could mislead a solver, but is still correct. \\
3. FP (False Positive): statements that are INCORRECT yet presented as if they were true. \\
4. FN (False Negative): statements that are INCORRECT because they deny or ignore facts that should be true. \\
\\
**Constraints** \\
- Do not mention ``Option A,'' ``the second choice,'' or ``the answer.'' \\
- Do not copy any words or phrases directly from the options list. \\
- Each proposition must be a single, complete sentence. \\
- Derive propositions solely from the underlying domain principles of the question stem, independent of the specific content or validity of the provided options. \\
- Return the propositions in a JSON array of strings.} \\

Target Type: \texttt{\{proposition\_type\}} \\
Question: \texttt{\{question\}} \\
Options: \texttt{\{options\}} \\

Please generate propositions.
    \end{promptbox}
    \end{adjustbox}
    
    \caption{\textbf{Optimized Proposition Prompt, $\phi_{\text{prop}}^{*}$, obtained by the DSPy Framework.} The gray-blue texts are iteratively updated (i.e., refined) during the joint prompt optimization. Note that $\{\texttt{proposition\_type}\}$ is filled with different values according to the target proposition category, resulting in a category-specific prompt. For example, if the target proposition category is True-Positive (TP), then $\{\texttt{proposition\_type}\}$ is replaced by \texttt{TP}. In other words, the value of ``Target Type'' in the prompt is filled with \texttt{TP} (i.e., Target Type: \texttt{TP}). As a result, the prompt becomes a TP-specific prompt, $\phi^{\text{TP}}_{\text{prop}}$, and by feeding this into the proposition model $\mathcal{M}_{\text{prop}}$, we can obtain up to $K$ propositions specific to True-Positive category, $\{ \hat{p}_{\text{TP}} \}_{k=1}^{K} \sim \mathcal{M}_{\text{prop}}(\phi^{\text{TP}}_{\text{prop}}, q)$. This process is repeated for all categories independently.}
    \label{fig:optimized-proposition-prompt}
\end{figure*}

% --- Figure 2: Answer Prompt ---
\begin{figure*}[ht!]
    \centering
    \begin{adjustbox}{width=1.0\textwidth}
    \begin{promptbox}{Optimized Prompt for $\mathcal{M}_{\text{ans}}$}
    \small 
    \renewcommand{\baselinestretch}{0.9}\selectfont
    
\textcolor{mutedblue}{You are a helpful AI assistant capable of complex reasoning. \\
First, analyze the question and think through the steps to solve it. \\
Throughout your analysis, strictly refer to the 'Proposition Definitions' to distinguish between essential facts and misconceptions. \\
Then, conclude your response with the final answer in the format `\#\#\# The answer is <only letter>`. \\
\\
Proposition Definitions: \\
1. [TP] : factual, essential knowledge that directly supports solving the problem. \\
2. [TN] : factual knowledge that reflects a common misconception or irrelevant fact that could mislead a solver, but is still correct. \\
3. [FP] : statements that are INCORRECT yet presented as if they were true. \\
4. [FN] : statements that are INCORRECT because they deny or ignore facts that should be true.} \\

\texttt{\{question\}} \\
\texttt{\{options\}} \\

Propositions:
\texttt{\{generated\_propositions\}} \\

Let's think step by step.
    \end{promptbox}
    \end{adjustbox}
    
    \caption{\textbf{Optimized Answer Prompt, $\phi_{\text{ans}}^{*}$, obtained by the DSPy Framework.} The gray-blue texts are iteratively updated (i.e., refined) during the joint optimization process. The input field \texttt{\{generated\_propositions\}} is filled with $\mathcal{P}$, the collection of concrete propositions generated by the proposition model $\mathcal{M}_{\text{prop}}$. To strictly comply with the original form of CoT prompt, we include \texttt{\{question\}} and \texttt{\{options\}} in the prompt without prepending their key names, such as ``Question:'' and ``Options:''. By feeding the prompt into the answer model $\mathcal{M}_{\text{ans}}$, the model generates a rationale and final answer based on the Chain-of-Thought trigger \textit{``Let's think step by step.''}}
    \label{fig:optimized-answer-prompt}
\end{figure*}

% --- Figure 3: Judge Prompt ---
\begin{figure*}[t]
    \centering
    \begin{adjustbox}{width=1.0\textwidth}
    \begin{promptbox}{Prompt for $\mathcal{M}_{\text{judge}}$}
    \small
    \renewcommand{\baselinestretch}{0.9}\selectfont
    
You are a meticulous logic evaluator for a QA system. \\
\\
1. TP Knowledge (True Positive Knowledge) refers to essential and specific knowledge points necessary for fundamentally understanding a problem. This includes concepts, theories, principles, laws, and factual information related to the problem's domain. Avoid overly broad categories; aim for detailed and precise knowledge points. \\
2. TN Knowledge (True Negative Knowledge) refers to indirect knowledge points that involve misconceptions or common mistakes that could lead to confusion or errors when solving a problem. This includes misconceptions, frequent errors, and incorrect information within the problem's domain. All knowledge points must be factually correct, and as with TP knowledge, they should be detailed and precise rather than overly general. \\
3. FP Knowledge (False Positive Knowledge) refers to factually false propositions asserted as correct. This covers fabricated or hallucinated claims, misapplied or over-generalized principles (e.g., using a theorem outside its conditions), outdated facts presented as current, and wrong attributions (entities, dates, units, causal links). \\
4. FN Knowledge (False Negative Knowledge) refers to knowledge points that are not factually correct and that wrongly deny, contradict, or express unjustified ignorance or uncertainty about facts that should be correct within the problem's scope. \\
    
Score the proposition's adherence to the definition from 0 to 1. \\

Question: \texttt{\{question\}} \\
Options: \texttt{\{options\}} \\
Correct Answer: \texttt{\{ground\_truth\_answer\}} \\
Generated Propositions: \texttt{\{generated\_propositions\}} \\
Target Type: \texttt{\{proposition\_type}\} \\
    \end{promptbox}
    \end{adjustbox}
    
    \caption{\textbf{Judge Prompt, $\phi_{\text{judge}}$, used by the Judge Model $\mathcal{M}_{\text{judge}}$.} This prompt defines the evaluation rubric for each proposition category (TP, TN, FP, FN). The model assesses whether the generated propositions adhere to the logical structure of their definitions. The output score, $s_{\text{quality}} \in [0, 1]$, becomes reward signal for DSPy optimizer.}
    \label{fig:judge-prompt}
\end{figure*}

\section{Details of Benchmark Datasets} \label{apdx: details-of-benchmark-datasets}

This section provides a comprehensive overview of the benchmark datasets employed to evaluate the efficacy of the concretized proposition prompting (CPP) method. To ensure a robust assessment across diverse reasoning axes---specifically targeting the balance between compositionality and knowledgeability---we selected eight distinct datasets spanning three primary domains: Commonsense, Math, and Medicine. Table \ref{tab:benchmark-stats} summarizes the key statistics of these datasets, including their domain categorization and the number of test samples used for evaluation.

\begin{table}[h]
    \centering
    \caption{Summary of benchmark datasets used in our experiments.}
    \label{tab:benchmark-stats}
    \resizebox{\columnwidth}{!}{%
    \begin{tabular}{l l c c}
    \toprule
    \textbf{Dataset} & \textbf{Domain} & \multicolumn{2}{c}{\textbf{Sample Size}} \\ 
    \cmidrule(lr){3-4}
    & & \textbf{Train} & \textbf{Test} \\ 
    \midrule
    ARC-E & Commonsense & 2,251 & 2,376 \\
    ARC-C & Commonsense & 1,119 & 1,172 \\
    MMLU-Pro & Commonsense & 0 & 12,032 \\ 
    CSQA & Commonsense & 9,741 & 1,221 \\ 
    \midrule
    GSM-8K & Math & 7,473 & 1,319 \\
    MATH & Math & 12,000 & 500 \\ 
    \midrule
    EHRNoteQA & Medicine & 0 & 962 \\ 
    MedXpertQA & Medicine & 0 & 2,450 \\ 
    \bottomrule
    \end{tabular}%
    }
\end{table}

\subsection{AI2 Reasoning Challenge (ARC)}
The AI2 Reasoning Challenge (ARC) dataset \citep{Clark2018ThinkYH} consists of 7,787 multiple-choice science questions from grades 3 through 9. It is divided into an Easy set (ARC-E) and a Challenge set (ARC-C). The ARC-C partition is specifically designed to include questions that retrieval-based algorithms and word co-occurrence models fail to answer, thus requiring more advanced reasoning logic. Most questions follow a four-option format, focusing on evaluating the model's ability to integrate disparate scientific facts.

\subsection{MMLU-Pro}
MMLU-Pro \citep{NEURIPS2024_ad236edc} is an enhanced version of the original MMLU, focusing on complex reasoning rather than simple knowledge retrieval. It addresses the issues of noise and easy retrieval in MMLU by filtering out trivial questions and introducing more reasoning-intensive benchmarks. A critical structural change is the expansion of the answer space from four to ten options per question, which significantly raises the difficulty ceiling and reduces the impact of random guessing on evaluation scores.

\subsection{CommonsenseQA (CSQA)}
CommonsenseQA \citep{talmor-etal-2019-commonsenseqa} is a large-scale multiple-choice dataset containing 12,102 questions. It is uniquely constructed using the ConceptNet knowledge graph to evaluate various types of commonsense reasoning (e.g., spatial, temporal, and social relations). Each question provides five answer options, where distractors are intentionally chosen to be semantically similar to the target concept, forcing models to distinguish between subtle commonsense nuances.

\subsection{GSM-8K}
GSM-8K \citep{cobbe2021} provides approximately 8.5K grade school math word problems that require multi-step reasoning. Each problem requires a sequence of 2 to 8 steps to reach a final numerical answer. Unlike multiple-choice datasets, GSM-8K uses an open-ended format where the model must generate a full solution trace, making it a standard benchmark for measuring a model's chain-of-thought and logical consistency in arithmetic tasks.

\subsection{MATH}
The MATH dataset \citep{hendrycks2021measuring} is a high-difficulty benchmark comprising 12,500 problems from middle and high school math competitions (e.g., AMC 10/12, AIME). It covers seven subjects: Algebra, Counting \& Probability, Geometry, Number Theory, Prealgebra, Precalculus, and Intermediate Algebra. Problems are presented in a free-response format requiring complex LaTeX-formatted step-by-step solutions, evaluating the model's ability to handle high-level abstract mathematical concepts.

\subsection{EHRNoteQA}
EHRNoteQA \citep{NEURIPS2024_e15c4aff} is a clinical reasoning benchmark built from real patient discharge summaries in the MIMIC-IV database. It consists of 962 expert-curated questions that require a deep understanding of longitudinal patient history and clinical notes. The dataset evaluates a model's clinical expertise and its ability to synthesize unstructured medical information for accurate diagnostic and treatment-related reasoning.

\subsection{MedXpertQA}
MedXpertQA \citep{zuo2025medxpertqa} is a specialized, expert-level benchmark designed to evaluate AI agents in diagnostic and clinical scenarios. It encompasses 12,710 high-stakes medical examination questions across diverse subspecialties. By prioritizing diagnostic reasoning over simple fact retrieval, MedXpertQA rigorously tests the knowledgeability axis, requiring models to demonstrate precision in both factual grounding and clinical decision-making.

\clearpage
\onecolumn
\section{Summary of Comparison Methods}

\begin{table}[ht!]
\centering
\scriptsize
\setlength{\tabcolsep}{5pt}
\renewcommand{\arraystretch}{1.5}

% NOTE:
% - Removed \resizebox to prevent the whole table from being scaled down.
% - Used tabularx with an X column so long text wraps vertically instead of shrinking.

\begin{tabularx}{\textwidth}{@{}p{0.15\textwidth} p{0.04\textwidth} p{0.26\textwidth} X@{}}
\toprule
\textbf{Method} & \textbf{Axis} & \textbf{Key Idea} & \textbf{Motivation} \\
\midrule

Plan-and-Solve \citep{wang-etal-2023-plan} & C &
Decompose into plan then execute solve following that plan. &
Despite the success of Zero-shot-CoT, it still suffers from three pitfalls: calculation errors, missing-step errors, and semantic misunderstanding errors. \\

Rephrase-and-Respond \citep{deng2023rephrase} & C &
Rephrase or expand the original question before answering. &
Misunderstandings arise not only in interpersonal communication but also
between humans and Large Language Models (LLMs). Such discrepancies can make LLMs interpret seemingly unambiguous questions in unexpected ways, yielding incorrect responses. \\

Thread-of-Thought \citep{zhou2023threadthoughtunravelingchaotic} & C &
Segment and iteratively analyze long/chaotic context with an explicit reasoning thread. &
LLMs encounter difficulties when confronted with chaotic contexts (e.g., distractors rather than long irrelevant context), leading to the inadvertent omission of certain details within the chaotic context. \\

Re-reading \citep{xu-etal-2024-reading} & C &
Reread the question/context to enhance understanding. &
Scant attention has been paid to the understanding of the input phase. \\

Meta-Prompting \citep{zhang2023meta} & C &
Use a meta-level prompt specifying formal problem structure instead of content exemplars. &
LLMs, in their typical operations, mirror System 1 processes and thus encounter difficulties with tasks that require the more deliberate, structured approach characteristic of System 2 thinking. \\

Least-to-Most \citep{zhou2022} & C &
Decompose complex problem into easy subproblems and solve sequentially. &
Chain-of-thought (CoT) often performs poorly on tasks which need to solve problems harder than the demonstration examples. \\

System-2-Attention \citep{weston20232attentionisneed} & K &
Regenerate input to retain only relevant information and then answer. &
Soft attention in Transformer-based Large Language Models (LLMs) is susceptible to incorporating irrelevant information from the context into its latent representations. \\

Analogical \citep{yasunaga2023} & K &
Generate analogical exemplars/knowledge relevant to the query, then answer. &
Chain-of-thought (CoT) prompting for language models demonstrates impressive performance across reasoning tasks, but typically needs labeled exemplars of the reasoning process. \\

SKE-Learn \citep{huang2025} & K &
Extract and verify internal knowledge, then use it in reasoning. &
SKE-Learn enables the model to quote explicit inner knowledge as evidence, thereby reducing hallucinations and enhancing overall quality of generated outputs. \\

\bottomrule
\end{tabularx}

\caption{\textbf{Classification of Comparison Methods along the Composition--Knowledge Dichotomy.} Motivations are excerpted from the abstract or introduction of each referenced paper. The letters C and K in the Axis column denote Compositionality and Knowledgeability, respectively.}
\label{tab4: classification-of-comparison-methods}
\end{table}

\clearpage
\onecolumn
\section{Examples of Final Outputs}

\begin{table}[ht!]
    \centering
    \small
    \renewcommand{\arraystretch}{1.2}
    
    \begin{tabularx}{\textwidth}{@{}X@{}}
        \toprule
        \textbf{Question:} Which life function is the same for an amoeba as for other freshwater protists? \\
        \textbf{Choices:} (A) the method used to obtain nutrients and energy \quad \textbf{(B) the process used to constantly remove excess water} \quad (C) the method used to detect changes in their environment \quad (D) the process used to remove waste products from their cytoplasm \\
        \midrule
        
        \textbf{Reasoning from Zero-shot CoT:} \\
        To determine which life function is the same for an amoeba as for other freshwater protists, we need to consider the basic biological processes that are common among these organisms. \newline
        A. The method used to obtain nutrients and energy: Amoebae and other freshwater protists can vary widely in how they obtain nutrients. Some may engulf food particles (phagocytosis), while others might absorb nutrients directly from the water or photosynthesize if they contain chloroplasts. This method can differ significantly between different species. \newline
        B. The process used to constantly remove excess water: Freshwater protists, including amoebae, face the challenge of osmotic pressure due to the hypotonic environment they live in. They generally have mechanisms like contractile vacuoles to expel excess water. This is a common feature among many freshwater protists. \newline
        C. The method used to detect changes in their environment: Sensory capabilities can vary greatly among protists. While some may have specialized structures to detect chemical or physical changes, others might not have such specific mechanisms. This is not a universally shared trait. \newline
        D. The process used to remove waste products from their cytoplasm: All living cells, including amoebae and other protists, need to remove waste products to maintain cellular homeostasis. This typically involves exocytosis or expulsion through cell membranes, which is a fundamental process across all these organisms. \newline
        Given this analysis, both B and D seem to be common functions. However, since the question asks for ``the'' life function, implying a singular answer, and considering the broader applicability of waste removal across all living cells, the most universally applicable answer would be D. \\
        \textbf{Output:} \textcolor{wrongred}{The answer is D.} \\
        \midrule
        
        \textbf{Reasoning from CPP:} \\
        \textbf{Propositions:} \newline
        [TP] Freshwater protists, including amoebae, use osmoregulation to manage the constant influx of water due to their hypotonic environment. \newline
        [TN] Amoebas, like many other freshwater protists, do not have a specialized method for detecting changes in their environment, often relying on simple physical responses instead. \newline
        [FP] Amoebas use the same method to obtain nutrients and energy as land plants do through photosynthesis. \newline
        [FN] Amoebas do not need to constantly remove excess water from their cells like other freshwater protists do. \newline
        \textbf{Reasoning process:} \newline
        To determine which life function is the same for an amoeba as for other freshwater protists, we need to evaluate each option against the provided propositions: \newline
        A. The method used to obtain nutrients and energy: This option is incorrect according to [FP0], which states that amoebas do not use the same method as land plants for obtaining nutrients and energy. However, [FP0] is incorrect itself, so this does not provide us with a direct answer. We know that amoebas typically engulf food particles through phagocytosis, which is a common method among many protists, but this is not explicitly stated as a shared trait in the given propositions. \newline
        B. The process used to constantly remove excess water: According to [TP0], freshwater protists, including amoebae, do indeed use osmoregulation to manage the constant influx of water due to their hypotonic environment. This suggests that this is a shared trait among these organisms. \newline
        C. The method used to detect changes in their environment: [TN0] indicates that amoebas do not have a specialized method for detecting environmental changes, similar to many other freshwater protists. While this proposition is factual, it does not highlight a shared life function but rather a lack thereof. \newline
        D. The process used to remove waste products from their cytoplasm: There is no specific proposition addressing this point directly. However, it is generally known that all living cells, including amoebas and other protists, must remove waste products to maintain cellular homeostasis. Since this is a fundamental life function, it can be inferred as a shared trait, but it is not explicitly supported by the given propositions. \newline
        Given the information provided, the most strongly supported shared life function based on the propositions is the process used to constantly remove excess water (osmoregulation). \\
        \textbf{Output:} \textcolor{correctblue}{The answer is B.} \\
        \bottomrule
    \end{tabularx}
    
    \caption{\textbf{Example of ARC-E.} Zero-shot CoT fails by over-generalizing a universal biological process (waste removal) while ignoring the specific ``freshwater'' context of the question. In contrast, CPP remains grounded in the habitat-specific requirement through its [TP] proposition regarding osmoregulation in hypotonic environments. This illustrates how CPP helps LLMs prioritize context-dependent facts over broad but less relevant knowledge.}
    \label{tab4: arc_e_full_text}
\end{table}

\begin{table}[ht!]
    \centering
    \small
    \renewcommand{\arraystretch}{1.2}
    
    \begin{tabularx}{\textwidth}{@{}X@{}}
        \toprule
        \textbf{Question:} A student sets up an experiment to learn which colors of light are refracted by a prism. Which method will the student most likely use to collect data? \\
        \textbf{Choices:} (A) research \quad (B) interviews \quad \textbf{(C) observation} \quad (D) measurement \\
        \midrule
        
        \textbf{Reasoning from Zero-shot CoT:} \\
        To determine which method a student would most likely use to collect data on which colors of light are refracted by a prism, let's consider each option: \newline
        A. Research: This involves gathering information from existing sources such as books or articles. While this can provide background knowledge, it doesn't directly involve collecting new data from the experiment itself. \newline
        B. Interviews: This involves asking questions to people who might have knowledge about the topic. This method is not suitable for collecting data on the physical properties of light and prisms. \newline
        C. Observation: This involves watching and noting what happens when light passes through the prism. This could be useful for seeing how different colors of light behave, but it may not provide precise details without additional tools. \newline
        D. Measurement: This involves using instruments to quantify the angles at which different colors of light are refracted. This method allows for accurate and repeatable data collection, which is crucial for understanding the behavior of light through a prism. \newline
        Given that the goal is to understand the refraction of light through a prism, the most appropriate method would be one that allows for precise data collection regarding the angles and behaviors of light. Therefore, the best choice is D. Measurement. \\
        \textbf{Output:} \textcolor{wrongred}{The answer is D.} \\
        \midrule
        
        \textbf{Reasoning from CPP:} \\
        \textbf{Propositions:} \newline
        [TP] The student will most likely use direct observation to collect data on the colors of light refracted by a prism. \newline
        [TN] The student might mistakenly believe that conducting interviews can provide useful data about light refraction, although this is not a scientific method for such an experiment. \newline
        [FP] The student will primarily use taste tests to determine which colors of light are refracted by a prism. \newline
        [FN] The student will not use observation to collect data on the refraction of light colors by a prism. \newline
        \textbf{Reasoning process:} \newline
        The task at hand involves understanding how a student would collect data on the refraction of light through a prism. Refraction is a physical phenomenon that can be observed and measured directly. \newline
        [TP0] correctly identifies that the primary method for collecting data in this scenario would be direct observation. When light passes through a prism, it separates into its component colors, creating a spectrum. This process can be visually observed and recorded. \newline
        [TN0] mentions a potential misconception where one might think interviews could provide useful data. However, interviews are not a scientific method for observing physical phenomena like light refraction; they are more suited for gathering qualitative information from individuals. \newline
        [FP0] incorrectly suggests that taste tests would be used to determine the colors of light refracted by a prism. This is nonsensical since taste has no relevance to the visual spectrum of light. \newline
        [FN0] incorrectly denies the use of observation, which is actually a crucial part of the experimental setup for studying light refraction. \newline
        Given these considerations, the most appropriate method for the student to collect data on the refraction of light by a prism is observation. \\
        \textbf{Output:} \textcolor{correctblue}{The answer is C.} \\
        \bottomrule
    \end{tabularx}
    
    \caption{\textbf{Example of ARC-C.} Zero-shot CoT fails by overthinking the task, mistakenly choosing ``measurement'' because it assumes a scientific experiment always requires technical precision. CPP remains grounded in the actual context through its [TP] proposition, correctly identifying that observing colors is a visual task. This demonstrates how CPP prevents LLMs from being distracted by sophisticated but contextually incorrect reasoning.}
    \label{tab5: arc_c_full_text}
\end{table}

\begin{table}[ht!]
    \centering
    \small
    \renewcommand{\arraystretch}{1.2}
    
    \begin{tabularx}{\textwidth}{@{}X@{}}
        \toprule
        \textbf{Question:} In one year, the number of students on campus doubles at the end of every month. If there are 10 students on campus at the beginning of the year, how many additional students would have joined by the end of May, above and beyond the number of students already on campus at the beginning of the year? \\
        \midrule
        
        \textbf{Reasoning from Zero-shot CoT:} \\
        To solve this problem, we need to calculate the number of students on campus at the end of each month from January to May, given that the number of students doubles at the end of every month. \newline
        1. \textbf{January}: The number of students is 10. \newline
        2. \textbf{February}: The number of students doubles, so \(10 \times 2 = 20\). \newline
        3. \textbf{March}: The number of students doubles again, so \(20 \times 2 = 40\). \newline
        4. \textbf{April}: The number of students doubles again, so \(40 \times 2 = 80\). \newline
        5. \textbf{May}: The number of students doubles again, so \(80 \times 2 = 160\). \newline
        
        Now, we need to find out how many additional students have joined by the end of May, compared to the beginning of the year. \newline
        - Number of students at the beginning of the year: 10 \newline
        - Number of students at the end of May: 160 \newline
        The additional students that have joined by the end of May are:
        \[
        160 - 10 = 150
        \]
        Therefore, the number of additional students that have joined by the end of May, above and beyond the number of students already on campus at the beginning of the year, is 150. \newline
        \textbf{Output:} \textcolor{wrongred}{150} \\
        \midrule
        
        \textbf{Reasoning from CPP:} \\
        \textbf{Propositions:} \newline
        [TP] By the end of May, the number of students on campus will have doubled 5 times, resulting in 320 students on campus. \newline
        [TP] The initial number of students on campus at the beginning of the year is 10. \newline
        [TP] To find the number of additional students by the end of May, subtract the initial number of students from the total number of students at the end of May. \newline
        [TN] The number of students doubles each month, which means the growth is exponential, not linear. \newline
        [FP] By the end of May, the number of students on campus will be 640, which means 630 additional students have joined since the beginning of the year. \newline
        [FN] The number of students does not double every month; it remains constant throughout the year. \newline
        
        \textbf{Reasoning process:} \newline
        To solve this problem, we need to follow these steps: \newline
        1. \textbf{Identify the initial number of students:} According to [TP1], there are 10 students on campus at the beginning of the year. \newline
        2. \textbf{Understand the doubling pattern:} The number of students doubles at the end of every month. This is an exponential growth pattern, as stated in [TN0]. \newline
        3. \textbf{Calculate the number of students at the end of each month:} \newline
        - End of January: \(10 \times 2 = 20\) \newline
        - End of February: \(20 \times 2 = 40\) \newline
        - End of March: \(40 \times 2 = 80\) \newline
        - End of April: \(80 \times 2 = 160\) \newline
        - End of May: \(160 \times 2 = 320\) \newline
        4. \textbf{Determine the number of additional students by the end of May:} According to [TP2], subtract the initial number from the total at end of May: \newline
        \[
        320 - 10 = 310
        \]
        Therefore, the number of additional students that would have joined by the end of May is 310. \newline
        \textbf{Output:} \textcolor{correctblue}{310} \\
        \bottomrule
    \end{tabularx}
    
    \caption{\textbf{Example of GSM-8K.} Zero-shot CoT makes a mistake by skipping the first doubling step at the end of January. It only calculates four doublings instead of five. CPP avoids this error by listing the doubling steps as a fact [TP] before doing the math. This list acts as a guide, ensuring the model calculates correctly every month from January to May.}
    \label{tab6: GSM-8K_full_text}
\end{table}

\begin{table}[ht!]
    \centering
    \small
    \renewcommand{\arraystretch}{1.2}
    
    \begin{tabularx}{\textwidth}{@{}X@{}}
        \toprule
        \textbf{Question:} One gear turns $33\frac{1}{3}$ times in a minute. Another gear turns 45 times in a minute. Initially, a mark on each gear is pointing due north. After how many seconds will the two gears next have both their marks pointing due north? \\
        \midrule
        
        \textbf{Reasoning from Zero-shot CoT:} \\
        To determine after how many seconds the two gears will next have both their marks pointing due north, we need to find the least common multiple (LCM) of the number of rotations each gear makes in a minute, and then convert that time into seconds. \newline
        
        1. \textbf{Determine the number of rotations per second for each gear:} \newline
        - The first gear turns \(33 \frac{1}{3}\) times in a minute. Converting this to an improper fraction:
        \[
        33 \frac{1}{3} = \frac{100}{3}
        \]
        Therefore, the first gear turns \(\frac{100}{3}\) times per minute. To find the number of rotations per second, we divide by 60 (since there are 60 seconds in a minute):
        \[
        \frac{100}{3} \div 60 = \frac{100}{3} \times \frac{1}{60} = \frac{100}{180} = \frac{5}{9}
        \]
        So, the first gear turns \(\frac{5}{9}\) times per second. \newline
        
        - The second gear turns 45 times in a minute. To find the number of rotations per second, we divide by 60:
        \[
        45 \div 60 = \frac{45}{60} = \frac{3}{4}
        \]
        So, the second gear turns \(\frac{3}{4}\) times per second. \newline
        
        2. \textbf{Find the LCM of the number of rotations per second:} \newline
        - We need to find the LCM of \(\frac{5}{9}\) and \(\frac{3}{4}\). First, express these as fractions with a common denominator:
        \[
        \frac{5}{9} = \frac{5 \times 4}{9 \times 4} = \frac{20}{36}
        \]
        \[
        \frac{3}{4} = \frac{3 \times 9}{4 \times 9} = \frac{27}{36}
        \]
        - Now, we need the LCM of the numerators (20 and 27) and the greatest common divisor (GCD) of the denominators (36). \newline
        
        3. \textbf{Calculate the LCM of the numerators:} \newline
        - The prime factorization of 20 is \(2^2 \times 5\). \newline
        - The prime factorization of 27 is \(3^3\). \newline
        - The LCM is:
        \[
        \text{LCM}(20, 27) = 2^2 \times 3^3 \times 5 = 540
        \]
        
        4. \textbf{Calculate the GCD of the denominators:} \newline
        - The prime factorization of 36 is \(2^2 \times 3^2\). \newline
        - \(\text{GCD}(36, 36) = 36\). \newline
        
        5. \textbf{Find the LCM of the fractions:} \newline
        \[
        \frac{\text{LCM}(20, 27)}{\text{GCD}(36, 36)} = \frac{540}{36} = 15
        \]
        
        6. \textbf{Convert the LCM from rotations per second to seconds:} \newline
        \[
        \text{Time in seconds} = \frac{15 \text{ rotations}}{\frac{5}{9} \text{ rotations/second}} = 15 \times \frac{9}{5} = 27 \text{ seconds}
        \]
        
        Therefore, the two gears will next have both their marks pointing due north after \(\boxed{27}\) seconds. \\
        \textbf{Output:} \textcolor{wrongred}{27} \\
        \bottomrule
    \end{tabularx}
    
    \caption{\textbf{Example of MATH.} Zero-shot CoT fails because it confuses the speed of the gears with the time needed to complete a full circle. It tries to find a common value for how fast the gears turn (rotations per second) instead of finding a common time when both gears finish a whole number of turns. This logical mix-up leads to the wrong final answer.}
    \label{tab7: math_full_text}
\end{table}

\begin{table}[ht!]
    \centering
    \small
    \renewcommand{\arraystretch}{1.2}
    
    \begin{tabularx}{\textwidth}{@{}X@{}}
        \toprule
        \textbf{Reasoning from CPP:} \\
        \textbf{Propositions:} \newline
        [TP] The least common multiple of the times it takes for each gear to complete a full rotation will determine when both marks point due north again. \newline
        [TN] The gears will align every least common multiple of their individual periods, which can be misleading if not calculated correctly. \newline
        [FP] The gears will align every 20 seconds, even though their rotation speeds are different. \newline
        [FN] The gears will never align their marks again after the initial position because their rotation speeds are not multiples of each other. \newline
        
        \textbf{Reasoning process:} \newline
        To solve this problem, we need to determine the time at which both gears will have completed an integer number of rotations and thus have their marks pointing due north again simultaneously. \newline
        
        Step 1: Convert the given rotations per minute into rotations per second. \newline
        - Gear 1 rotates \(33 \frac{1}{3}\) times per minute, which is equivalent to \(\frac{100}{3}\) rotations per minute. \newline
        - Therefore, in one second, Gear 1 rotates \(\frac{100}{3} \div 60 = \frac{100}{180} = \frac{5}{9}\) rotations. \newline
        - Gear 2 rotates 45 times per minute. \newline
        - Therefore, in one second, Gear 2 rotates \(45 \div 60 = \frac{3}{4}\) rotations. \newline
        
        Step 2: Determine the time in seconds when both gears complete an integer number of rotations. \newline
        - For Gear 1, the time \(t\) in seconds when it completes an integer number of rotations is \(t = \frac{9k}{5}\) where \(k\) is an integer. \newline
        - For Gear 2, the time \(t\) in seconds when it completes an integer number of rotations is \(t = \frac{4m}{3}\) where \(m\) is an integer. \newline
        
        Step 3: Find the least common time \(t\) such that both conditions are satisfied. \newline
        - Set \(\frac{9k}{5} = \frac{4m}{3}\). Cross-multiplying gives \(27k = 20m\). \newline
        - The smallest positive solution occurs when \(k\) is a multiple of 20 and \(m\) is a multiple of 27. \newline
        
        Step 4: Substitute back to compute \(t\). \newline
        - Using \(k = 20\): \(t = \frac{9 \times 20}{5} = 36\) seconds. \newline
        - Check with \(m = 27\): \(t = \frac{4 \times 27}{3} = 36\) seconds. \newline
        
        Thus, the two gears will next have both their marks pointing due north after 36 seconds. \newline
        \textbf{Output:} \textcolor{correctblue}{36} \\
        \bottomrule
    \end{tabularx}
    
    \caption{\textbf{Example of MATH (Continued).} CPP succeeds by first defining a clear goal in its [TP] propositions: finding the exact time when both gears finish full rotations at the same moment. By focusing on the time required for one full turn (the period) instead of just the speed, CPP follows a simple and correct logical path to reach the answer of 36 seconds.}
    \label{tab7: math_full_text (cont.)}
\end{table}

\begin{table}[ht!]
    \centering
    \small
    \renewcommand{\arraystretch}{1.2}
    
    \begin{tabularx}{\textwidth}{@{}X@{}}
        \toprule
        \textbf{Question:} Was the patient's pregnancy full-term during her second recorded admission, and why was a cesarean section required? \\
        \textbf{Choices:}
        (A) The admission was due to a kidney stone issue, and she underwent cesarean because of a repeat procedure. \quad
        (B) The pregnancy was not full-term. The cesarean was conducted because of non-reassuring fetal monitoring. \quad
        \textbf{(C) The patient's pregnancy was a full-term. The cesarean section was required due to non-reassuring fetal monitoring.} \\
        (D) The admission was for an oropharyngeal mass, and the cesarean section was performed due to previous such surgeries. \quad
        (E) The pregnancy was not full-term, and the cesarean section was performed as a routine procedure. \\
        \midrule
        
        \textbf{Reasoning from Zero-shot CoT:} \\
        To determine the correct answer, we need to break down the information provided in each option and evaluate it against the criteria given in the question: whether the pregnancy was full-term and the reason for the cesarean section. \newline

        1. \textbf{Option A:} Kidney stone issue + repeat procedure; does not clarify term status or specific reason. \newline
        2. \textbf{Option B:} Not full-term + non-reassuring fetal monitoring; provides both required parts. \newline
        3. \textbf{Option C:} Full-term + non-reassuring fetal monitoring; provides both required parts but contradicts B on term status. \newline
        4. \textbf{Option D:} Oropharyngeal mass + previous surgeries; unrelated to term status/NRFM. \newline
        5. \textbf{Option E:} Not full-term + routine; lacks medical necessity. \newline

        Given these analyses, the answer chosen is B. \\
        \textbf{Output:} \textcolor{wrongred}{The answer is B.} \\
        \midrule
        
        \textbf{Reasoning from CPP:} \\
        \textbf{Propositions:} \newline
        [TP] The patient's pregnancy was full-term at the time of her second recorded admission. \newline
        [TP] Non-reassuring fetal monitoring was the primary reason for the cesarean section. \newline
        [TN] Non-reassuring fetal monitoring can be a valid reason for performing a cesarean section, even if the pregnancy is not full-term. \newline
        [FP] The patient's pregnancy was not full-term, but the cesarean section was required due to a kidney stone issue causing complications. \newline
        [FN] The patient's pregnancy was not full-term, but the cesarean section was not required due to non-reassuring fetal monitoring. \newline

        \textbf{Reasoning process:} \newline
        [TP0] states the pregnancy \emph{was full-term}, and [TP1] states the C/S was due to \emph{non-reassuring fetal monitoring}. \newline
        Among the choices, only (C) matches \textbf{both} full-term status and non-reassuring fetal monitoring as the indication. \newline
        Therefore, CPP selects option C. \\
        \textbf{Output:} \textcolor{correctblue}{The answer is C.} \\
        \bottomrule
    \end{tabularx}
    
    \caption{\textbf{Example of EHRNoteQA.} Zero-shot CoT fails by picking a half-correct answer. It correctly identifies the reason for the surgery but makes a mistake about whether the pregnancy was full-term. CPP avoids this by breaking the question into two separate facts (propositions). By checking both the pregnancy term and the surgery reason separately, CPP ensures the model selects the only option that matches all the medical details correctly.}
    \label{tab8: ehrnoteqa_full_text_521}
\end{table}

\clearpage
\section{Failure Case Study}

\begin{table}[ht!]
    \centering
    \small
    \renewcommand{\arraystretch}{1.2}
    
    \begin{tabularx}{\textwidth}{@{}X@{}}
        \toprule
        \textbf{Question:} Solve for $x$: $3^{2x} + 19 = 10^x$. \\
        \midrule
                
        \textbf{Reasoning from CPP:} \\
        \textbf{Propositions:} \newline
        [TP] To solve the equation, you can use logarithms to bring down the exponents and isolate $x$. \newline
        [TN] The equation can be solved by taking the logarithm of both sides, but this does not necessarily simplify the problem into a standard algebraic form. \newline
        [FP] The equation can be solved by taking the logarithm of both sides and then applying the quadratic formula to find $x$. \newline
        [FN] The equation $3^{2x} + 19 = 10^x$ has no real solutions because the left side is always greater than the right side for all real values of $x$. \newline

        \textbf{Reasoning process:} \newline
To solve the equation \(3^{2x} + 19 = 10^x\), let's carefully analyze the propositions and proceed step-by-step: \\
1. \textbf{Understanding the Equation}:\\
- We have \(3^{2x} + 19 = 10^x\). \\
- This is an exponential equation where the terms involve different bases (3 and 10). \\

2. \textbf{Applying Logarithms}: \\
- [TP0]: Using logarithms can help us bring down the exponents, but it might not immediately simplify the equation into a standard algebraic form. However, it is a valid approach. \\
- [TN0]: Taking the logarithm of both sides is a valid method, but it doesn't guarantee simplification into a standard algebraic form. It can be useful for further analysis. \\
- [FP0]: Applying the quadratic formula directly after taking logarithms is incorrect because the equation is not in a quadratic form after logarithmic transformation. \\
- [FN0]: This proposition suggests there are no real solutions, which we need to verify. \\

3. \textbf{Analyzing the Propositions}: \\
- [TP0] and [TN0] are correct in their respective contexts. \\
- [FP0] is incorrect; the quadratic formula cannot be applied directly here. \\
- [FN0] requires verification. \\

4. \textbf{Verification of No Real Solutions}: \\
- Let's consider the behavior of the functions involved: \\
- \(3^{2x}\) grows exponentially as \(x\) increases. \\
- \(10^x\) also grows exponentially as \(x\) increases. \\
- The constant term 19 adds a fixed value to \(3^{2x}\). \\
- For small values of \(x\): - When \(x = 0\), \(3^{2 \cdot 0} + 19 = 1 + 19 = 20\) and \(10^0 = 1\). Clearly, \(20 > 1\). \\
- As \(x\) increases, \(3^{2x}\) grows much faster than \(10^x\), making \(3^{2x} + 19\) significantly larger than \(10^x\). \\
5. \textbf{Conclusion}: \\
- \textcolor{wrongred}{Given the rapid growth of the exponential terms, it is clear that \(3^{2x} + 19\) will always be greater than \(10^x\) for any real value of \(x\).} \\
- Therefore, the equation \(3^{2x} + 19 = 10^x\) has no real solutions. \\ \textbf{the answer is} No real solutions \\
\textbf{Output:} No real solutions. \\
        \bottomrule
    \end{tabularx}
    
\caption{\textbf{MATH Failure.} While CPP correctly identifies [TP] and [TN] as valid approaches and rejects [FP], it fails in the final verification step. The red-colored sentence indicates the critical reasoning error where the model incorrectly assumes that the left-hand side ($3^{2x} + 19$) is always greater than the right-hand side ($10^x$). This flawed intuition regarding exponential growth leads to the validation of the incorrect proposition [FN], causing the model to conclude ``no real solutions'' and miss the actual solution $x = 2$.}

    \label{tab:math_failure_83}
\end{table}

\begin{table}[ht!]
    \centering
    \small
    \renewcommand{\arraystretch}{1.2}
    
    \begin{tabularx}{\textwidth}{@{}X@{}}
        \toprule
        \textbf{Question:} What procedure was performed to manage the patient's severe pre-eclampsia during her last delivery? \\
        \textbf{Choices:} (A) The patient was administered with labetalol in IV form. 
        (B) For her wound cellulitis, she was started on Keflex.
        (C) The patient was given magnesium sulfate.
        \textbf{(D) The patient received a cesarean section along with bilateral tubal ligation.}
        (E) The patient was treated with nifedipine 30mg CR QD. \\
        \midrule
                
        \textbf{Reasoning from CPP:} \\
        \textbf{Propositions:} \newline
        [TP] Magnesium sulfate is commonly used to prevent seizures in patients with severe pre-eclampsia. \newline
        [TN] Labetalol is commonly used to manage hypertension in pregnant women, but it is not the only option available. \newline
        [FP] The patient was treated with high-dose aspirin to manage her severe pre-eclampsia during her last delivery. \newline
        [FN] The patient did not receive any antihypertensive medication during her last delivery for managing severe pre-eclampsia. \newline

        \textbf{Reasoning process:} \newline
To determine the correct procedure performed to manage the patient's severe pre-eclampsia during her last delivery, let's analyze each proposition and the options provided. \\
1. \textbf{Option A: The patient was administered with labetalol in IV form.} \\
   - This option aligns with [TN], which states that labetalol is commonly used to manage hypertension in pregnant women. However, we need to verify if it's the most appropriate choice for severe pre-eclampsia. \\
2. \textbf{Option B: For her wound cellulitis, she was started on Keflex.} \\
   - This option seems unrelated to managing severe pre-eclampsia. It might be relevant to the patient's overall care but doesn't address the specific condition in question. \\
3. \textbf{Option C: The patient was given magnesium sulfate.} \\
   - This option directly aligns with [TP], which states that magnesium sulfate is commonly used to prevent seizures in patients with severe pre-eclampsia. This makes it a strong candidate for the correct answer. \\
4. \textbf{Option D: The patient received a cesarean section along with bilateral tubal ligation.} \\
   - While this option might be relevant to the patient's delivery, it doesn't specifically address the management of severe pre-eclampsia. \\
5. \textbf{Option E: The patient was treated with nifedipine 30mg CR QD.} \\
   - Although nifedipine can be used to manage hypertension, there's no direct information linking it to the management of severe pre-eclampsia in this context. \\
\textcolor{wrongred}{Considering the information provided and the relevance to managing severe pre-eclampsia, \textbf{Option C} is the most accurate choice based on [TP].} \\
\textbf{the answer is C} \\
\textbf{Output:} C. \\
        \bottomrule
    \end{tabularx}
    
\caption{\textbf{EHRNoteQA Failure.} The question specifically asks for the procedure performed during the patient's last delivery. However, the model commits a category error by selecting magnesium sulfate, a medication, instead of the correct delivery procedure described in label Option D. The pipeline erroneously maps a true proposition regarding magnesium usage directly to the answer, ignoring the semantic distinction between a procedure and a medication. The red text highlights the specific step where this mismatch occurs and the model locks in the incorrect prediction.}
    \label{tab:ehr_failure_512}
\end{table}

\begin{table}[ht!]
    \centering
    \small
    \renewcommand{\arraystretch}{1.2}
    
    \begin{tabularx}{\textwidth}{@{}X@{}}
        \toprule
        \textbf{Question:} What was the patient's alcoholism status during her second admission to the hospital? \\
        \textbf{Choices:} \textbf{(A) Her alcoholism status was sober for several years.} (B) She was intoxicated. (C) She drank a 2L bottle of wine and a pint of vodka which resulted in coffee ground emesis. (D) She refused AMA forms and left the hospital. (E) Her alcoholism status was sober for several years., but she drank alcohol once 5 days ago \\
        \midrule
                
        \textbf{Reasoning from CPP:} \\
        \textbf{Propositions:} \newline
        \textcolor{wrongred}{[TP] The patient's alcoholism status during her second admission indicated recent alcohol consumption.} \newline
        [TN] Patients who have been sober for several years can sometimes relapse, but this does not necessarily mean they are consistently intoxicated. \newline
        [FP] The patient had been completely abstinent from alcohol for over a decade before her second admission. \newline
        [FN] The patient had been completely abstinent from alcohol for over a decade before her second admission. \newline

        \textbf{Reasoning process:} \newline
In this methodology, \emph{TP/TN should be true propositions grounded in the note}. This failure is driven by an incorrect (or at least unsupported) TP that injects a false constraint into answer selection. \\
1. \textbf{Check what the question asks: ``alcoholism status during second admission.''} \\
   - The answer choices are largely \emph{status labels} (A/E) vs. \emph{events} (B/C/D). \\
   - The gold answer is A: \emph{``sober for several years''}. \\
2. \textbf{Where the reasoning goes wrong: using TP as a hard filter.} \\
   - The model assumes the status must reflect \emph{recent drinking} because [TP] claims that the second admission indicates recent alcohol consumption. \\
   - This makes A look incompatible and pushes the model toward E. \\
3. \textbf{Why this implies a faulty proposition.} \\
   - If the record (and label) says the alcoholism status is ``sober for several years'' during the second admission, then a TP asserting ``recent alcohol consumption'' at that admission is inconsistent with the labeled status. \\
   - Therefore, the most plausible explanation is that \textcolor{wrongred}{the TP is not actually true/grounded for the second admission} (e.g., it may belong to a different admission, be a misread temporal cue, or confuse history-of-relapse with current status). \\
\textcolor{wrongred}{Because the (supposed) TP incorrectly states that the second admission involved recent alcohol consumption, the model treats Option A as impossible and selects Option E as the only choice that satisfies the erroneous TP cue.} \\
\textbf{the answer is E} \\
\textbf{Output:} E. \\
\bottomrule
    \end{tabularx}
    
\caption{\textbf{EHRNoteQA Failure.} The CPP operates on the premise that TP and TN propositions are true and grounded in the notes. Here, the TP proposition incorrectly claims recent alcohol consumption, contradicting the gold label which describes the patient as sober for several years. This error introduces a false constraint that forces the model to reject the correct label Option A and select Option E instead. The red text marks the specific point where the CPP's validity assumption fails and propagates into the final wrong prediction.}
    \label{tab:ehr_failure_55_faultyTP}
\end{table}

\end{document}